\documentclass[]{article}

\usepackage{amsfonts}
\usepackage{float}
\usepackage{amsmath}
\usepackage{multicol}
\usepackage{multirow}
\usepackage{amssymb}
\usepackage{graphicx}
\usepackage{subcaption}
\usepackage{caption}
\usepackage{float}
\usepackage{tabularx}
\usepackage{fontsize}
\usepackage{diagbox}
\usepackage{algorithm}
\usepackage{algpseudocode}
\usepackage{makecell}
\usepackage{longtable}
\usepackage{authblk}
\usepackage{hyperref}
\usepackage{xcolor}
\usepackage[sort]{natbib}

\setcitestyle{numbers,open={[[},close={]]}}
\title{Assessing Reliability of BERT-Based Models on Question Answering Tasks}

\author[a]{Pooja Yadav\thanks{Email: poojayadav234422@gmail.com }}
\author[a]{Priyanka Harjule}
\author[b]{Basant Agarwal}
\author[c]{Marko Robnik Šikonja\thanks{Corresponding author. Email: Marko.RobnikSikonja@fri.uni-lj.si}}
\affil[a]{Department of Mathematics, Malaviya National Institute of Technology, Jaipur, Rajasthan, India}
\affil[b]{Department of Computer Science and Engineering, Central University of Rajasthan, Kishangarh, Rajasthan, India}
\affil[c]{University of Ljubljana, Faculty of Computer and Information Science, Ljubljana, Slovenia}

\date{}

\begin{document}

\maketitle

\textbf{Abstract:} Reliability estimation of large language models is in many cases as crucial as their accuracy, as reliable models are more trustworthy, robust, and suitable for practical applications. Recent advancements in natural language processing (NLP), particularly those based on transformer architectures, have significantly accelerated progress across various NLP tasks. This study focuses on the reliability of transformer-based question answering (QA) models, specifically  BERT models and its variants (RoBERTa, ALBERT, DistilBERT). These encoder-only pretrained transformers have demonstrated remarkable accuracy in QA tasks that can be treated as classification tasks. However, their reliability remains underexplored. This study evaluates the reliability of four BERT-based models by assessing response stability under two conditions: (1) internal model variations induced via Monte Carlo Dropout (MCD) and (2) input perturbations through paraphrasing. Using the SQuAD and QuAC datasets, we investigate how dropout rates affect prediction consistency and whether lexical changes impact answer stability. Our findings reveal that RoBERTa maintains higher reliability, whereas AlBERT and DistilBERT exhibit significant inconsistencies. Statistical analyses confirm that enabling MCD during prediction does not disrupt inference dynamics, validating its effectiveness as a reliability metric. These findings underscore the importance of evaluating both accuracy and stability in QA models to ensure stability in real-world applications. \\


\section{Introduction}
\label{int}
In recent years, the field of Natural Language Processing (NLP) has witnessed remarkable advancements. Rapid advances in Natural Language Processing (NLP) have been transformative, driving significant progress in the development of intelligent systems capable of processing, understanding, and generating human language with remarkable accuracy. These advancements have not only enhanced the efficiency of language-based applications but have also expanded the scope of tasks that machines can handle, from basic text processing to complex conversation. Central to this progress is the introduction of transformer-based architectures, which have redefined the capabilities of machine learning models by employing self-attention mechanisms \cite{5} to capture complex linguistic patterns and contextual dependencies effectively. These architectures, characterized by their scalability and adaptability, have set new benchmarks across a wide range of NLP tasks. Moreover, the robustness and versatility of transformer-based models have made them indispensable tools for advancing research and practical applications, such as healthcare, customer support, and information retrieval. Their ability to model bidirectional context and manage long-range dependencies has enabled breakthroughs in many NLP tasks \cite{20}. \par
Among diverse tasks, question answering (QA) has emerged as a critical benchmark for evaluating language models' comprehension and reasoning capabilities. Transformer-based representational architectures, such as BERT-based models, have achieved remarkable accuracy across various datasets. QA systems are crucial for applications such as virtual assistants, customer support systems, and automated knowledge retrieval. Several authors \cite{15, 16} have concentrated on enhancing performance with architectural innovations, pre-training strategies, and various optimization techniques. Given their relatively good computational efficiency and impressive performance, BERT-like models are appropriate candidates for this task in many circumstances.   \par
However, while traditional evaluation metrics provide a quantitative measure of performance, they often present an incomplete picture of the model's capabilities. Metrics like accuracy primarily reflect how well models perform under idealized conditions, but they fail to capture how models react to uncertainty, internal stochastic variations, or input modifications. Since real-world applications often involve unpredictable conditions, it is crucial to assess how models respond to such changes to ensure reliability and consistency. The motivation for this study arises from the observation that, despite their high accuracy, the reliability and prediction stability of BERT-based QA models under controlled variations have not been systematically explored. \par
This study distinguishes itself through its methodological framework. It introduces a reliability assessment methodology that systematically examines the consistency and stability of BERT variants under stochastic variations and input perturbations. This analysis investigates the impact of two key perturbations -- internal stochastic variations and input modifications -- to identify potential weaknesses in model behavior. The considered models leverage transformer-based architectures to deliver high accuracy by effectively capturing linguistic nuances and context. BERT models have demonstrated consistent performance in various tasks, including QA \cite{26}. In 2022, \citet{7} established that BERT can be comparatively reliable in classification tasks. \citet{27} demonstrated that RoBERTa outperformed BERT with optimized training, achieving higher accuracy on the SQuAD and GLUE benchmarks. \citet{28} observed that ALBERT achieved comparable or better performance than BERT on QA tasks with fewer parameters. While their performance in terms of accuracy has been evaluated and documented in the literature using various datasets, such as SQuAD, TriviaQA, WikiQA, and QuAC \cite{1,3}, there remains a need to evaluate their reliability. This study seeks to address this gap by examining whether BERT models and their variants, which perform consistently in QA tasks, are equally reliable under varying conditions.   \par
Reliability, in the context of question answering, refers to the consistency of a model in generating answers to input queries when subjected to controlled perturbations either in the model's configuration or input. In the context of NLP, a change in input refers to when the input undergoes modifications, such as the use of alternate phrases, synonyms, or added noise. This process assesses the model’s reliability by examining its stability under controlled variations and evaluating the variability in outputs produced. In this study, we examine this concept by examining the model's reliability through controlled changes to its internal configuration and input variation for BERT-based QA models and analyzing the variation in outputs generated. This broader evaluation offers a deeper understanding of the model's stability and dependability. Reliability is quantified by analyzing the variance in outputs generated by a model when small alterations are made to the model itself or in the input provided, ensuring a comprehensive evaluation of its stability. This study primarily focuses on: 
\begin{enumerate}
    \item To evaluate the reliability of four BERT model variants—RoBERTa, BERT-Base, DistilBERT, and ALBERT—in question-answering tasks.
    \item To investigate how changes in model configuration and input variations affect the consistency of the generated answers.
    \item Investigating the relationship between model accuracy and reliability using two datasets, SQuAD (high accuracy) and QuAC (lower accuracy).
    \item To provide insights into the stability of BERT variants, contributing to the development of more stable NLP systems.
\end{enumerate}
This paper is organized into five sections. Section \ref{rel} presents a review of related work in the field. Section \ref{met} outlines the research methodology, including detailed information about the datasets utilized in the study. Section \ref{res} reports the results obtained from the analysis, followed by Section \ref{err}, which provides a brief analysis of instances where the models fail to answer. Finally, Section \ref{dis} provides an in-depth discussion of these findings, and Section \ref{con} concludes the study, summarizing the key insights and implications of the research.

\section{Related Work}
\label{rel}
Rawat and Samant \cite{2} conducted a detailed analysis of transformer-based models for question answering, focusing on BERT, ALBERT, RoBERTa, XLNet, DistilBERT, Electra, and Pegasus. Using the SQuAD2 dataset, they demonstrated how transformer models, such as HuggingFace's BERT QA model, outperform traditional "Bag of Words" approaches in extracting answers from large documents. Van Aken et al. \cite{14} examine various LLMs, including BERT, fine-tuned for Question Answering (QA), to explore how they transform token vectors to find correct answers, with a layer-wise evaluation of hidden states to extract valuable information. Nassiri and Akhloufi \cite{4} presented a comprehensive review of transformer models in text-based QA systems, categorizing architectures into encoders, decoders, and encoder-decoders. Their study also highlighted trends in QA datasets, system architectures, and evaluation methods, stressing the importance of simplifying transformer model implementations. Ozkurt \cite{1} explored the strengths of various transformer-based models for question answering and identified ALBERT as the top performer, achieving an impressive 86.85\% exact match and 89.91\% $F_1$ score on the SQuAD v2 dataset. Similarly, Kate Pearce and Tiffany Zhan \cite{3} conducted a comprehensive study of transformer models across diverse QA datasets. Their findings revealed that RoBERTa and BART pre-trained models consistently outperformed others, while their custom BERT-BiLSTM model surpassed the baseline BERT model in $F_1$ score.    \par
Despite the superior accuracy of these models, their reliability remains underexplored. The existing literature explores diverse approaches for assessing the reliability of language models \cite{6,8,11}, with methodologies adapting to specific problem domains. Miok et al. \cite{9} highlighted the effectiveness of LSTM models with Monte Carlo Dropout in enhancing reliability in the classification tasks (hate speech classification). In a separate study, Miok et al. \cite{7} evaluated the reliability of LSTM and BERT variants with Monte Carlo Dropout (MCD BERT). They found that MCD BERT variants, including BERT-base and mBERT, are particularly reliable for classification. \par
Previous studies have extensively highlighted that various BERT model variants can be designed and optimized for different Natural Language Processing (NLP) tasks, showcasing their versatility and adaptability across a wide range of real-world applications. Among these, RoBERTa, BERT-Base, DistilBERT, and ALBERT have been identified as widely adopted models in question-answering (QA) tasks \cite{15, 16}. \par
Recent studies have also explored the use of Large Language Models (LLMs) in question answering and recommendation-oriented applications. LLM-based recommendation frameworks such as LE-DLCM have shown that large language models can improve semantic understanding and contextual reasoning in personalized recommendation tasks \cite{ma2025dlcm}. \cite{wu2024survey, lin2025can} have similarly investigated conversational recommendation and generative reasoning using LLMs, mainly focusing on improving recommendation quality, user interaction, and generative capabilities of language models. \cite{shehmir2025llm4rec} highlighted the growing integration of LLMs into recommendation pipelines for semantic representation learning, conversational interaction, and context-aware recommendation strategies. These developments indicate the increasing role of LLM-driven semantic reasoning in modern NLP-based recommendation and QA-related systems, while also motivating the need for systematic reliability evaluation frameworks for question-answering models.

\section{Experimentation and Methodology}
\label{met}
This section outlines the datasets used in this study and the methodology employed to evaluate the reliability of the selected models. To comprehensively evaluate reliability, two approaches were incorporated into the analysis: one involving the assessment of reliability by inducing changes in the model's internal configuration, and the other by introducing variations in the input.    \par 
To examine reliability under internal configuration changes while keeping the input constant, stochasticity was introduced in the model’s outputs during the prediction phase, allowing an assessment of whether the generated answers remained semantically similar to the correct answers. This helps in evaluating the model’s reliability. In addition, reliability was evaluated by modifying the input while maintaining a constant model configuration to assess whether models prioritize syntactic structure over semantic comprehension. Specifically, a pre-trained paraphrasing model is used to generate a paraphrased version of the input, and the impact on the model’s responses was observed. This analysis facilitated the determination of whether the model maintained answer consistency despite minor lexical changes. A detailed explanation of the datasets and the methodology employed is provided in the following sections. \par
In this work, reliability refers to the consistency of model predictions under controlled perturbations. The proposed framework evaluates reliability by examining whether transformer-based question-answering models continue to generate semantically consistent answers when variations are introduced either within the model through Monte Carlo Dropout or at the input level through paraphrased questions. The resulting variations in the generated outputs are analyzed to assess the consistency and stability of the model’s prediction behaviour under controlled perturbation settings.

\subsection{Data Description}
\begin{figure}[H]
    \centering
    \includegraphics[width= \textwidth, keepaspectratio]{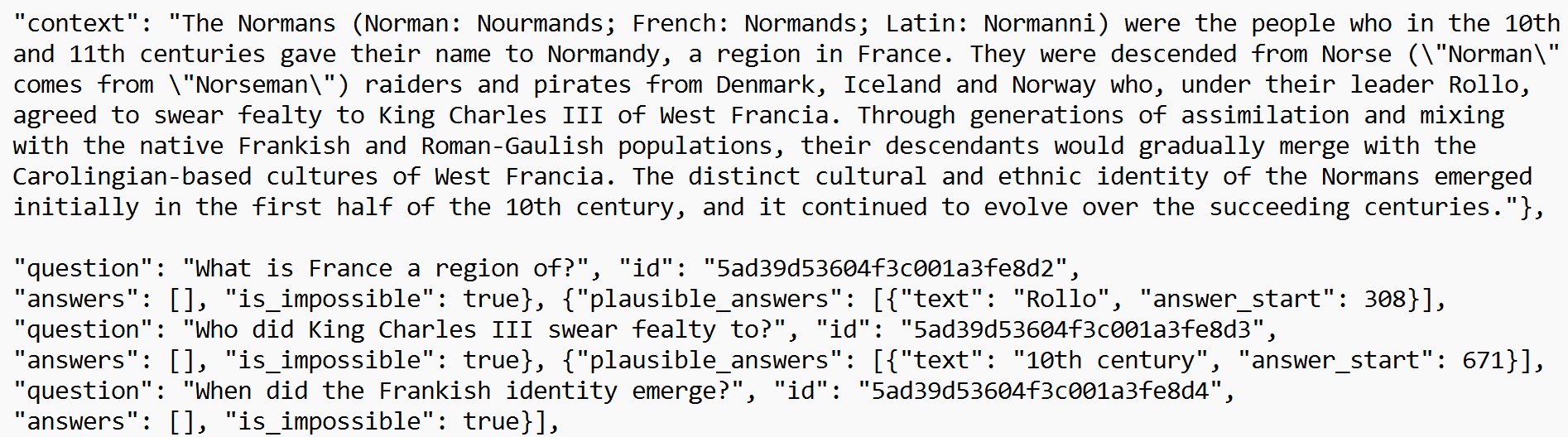}
    \caption{Visual Representation of the SQuAD Dataset.}
    \label{fi1}
\end{figure}

\begin{figure}[H]
    \centering
    \includegraphics[width= \textwidth, keepaspectratio]{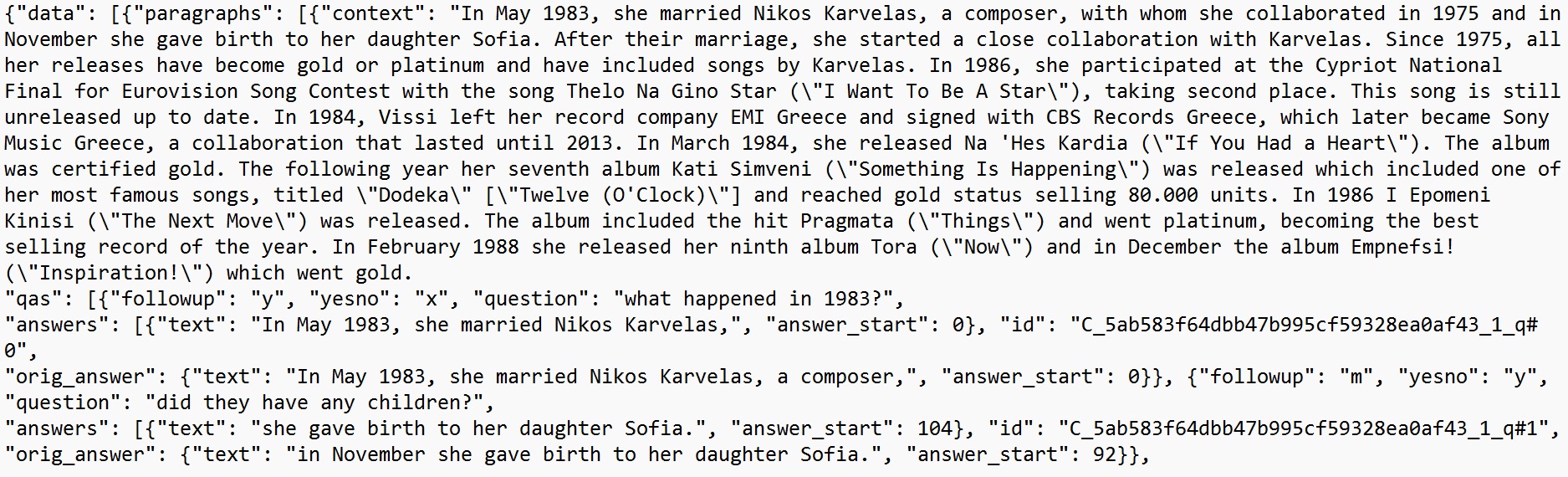}
    \caption{Visual Representation of the QuAC Dataset.}
    \label{fi2}
\end{figure}

To evaluate the reliability of the BERT models in question-answering tasks, two datasets were selected: the Stanford Question Answering Dataset (SQuAD) 2.0 and the Question Answering in Context (QuAC) dataset. Both datasets used in this study are publicly available. The SQuAD 2.0 dataset is available at \href{https://rajpurkar.github.io/SQuAD-explorer/}{SQuAD}, and the QuAC dataset is available at \href{https://quac.ai/}{QuAC}. Both datasets are widely used for question-answering tasks but differ significantly in design and purpose. SQuAD 2.0 features independent fact-based questions, enabling BERT models to perform consistently. In contrast, QuAC is designed for conversational QA, where the questions are context-dependent, forming a dialogue that requires understanding of previous interactions, leading to comparatively lower performance for BERT variants. Both data sets consist of a given context, corresponding questions related to that context, and the correct answer extracted from the provided passage. The context refers to the passage that serves as the source of information from which the answers to the questions must be derived. Additionally, questions for which the context lacks sufficient information to derive an answer are categorized as unanswerable.   
\begin{table}[H]
    \centering
    \begin{tabular}{|c|c|c|c|c|} \hline
 \textbf{Dataset}  & \textbf{Context} & \textbf{Total Questions} & \textbf{with Ans} & \textbf{No Answer}  \\  \hline
 SQuAD & 1204 & 11873 & 5928 & 5945  \\  \hline     QuAC & 1000 & 7354 & 5868 & 1486 \\   \hline    \end{tabular}
\caption{Description of used QA Datasets.}
 \label{aa}
\end{table}
The SQuAD 2.0 dataset \cite{12} includes both answerable and unanswerable questions. This is a benchmark dataset, commonly used to evaluate model performance. BERT and its variants have demonstrated remarkable accuracy on this dataset, making it a standard reference for evaluation. In contrast, the second dataset, QuAC \cite{13}, presents stronger challenges for BERT models, as they tend to exhibit significantly lower accuracy when evaluated on it. The primary motivation for selecting these two datasets is to investigate whether a model’s reliability is correlated with its accuracy. Specifically, the objective is to determine whether models that achieve high accuracy also demonstrate greater reliability or whether reliability is independent of accuracy. A summary of the SQuAD and QuAC datasets is provided in Table \ref{aa}. Both datasets are publicly available. Figures \ref{fi1} and \ref{fi2} show visual snapshots of the SQuAD and QuAC datasets, highlighting the context, questions, and answers.

\subsection{Architectural Details of Models}
This study employs four BERT variants—BERT-Base, RoBERTa, ALBERT, and DistilBERT—each derived from the original BERT model but with unique architectural adjustments to enhance performance, efficiency, or scalability. The following section provides a brief overview of these models, highlighting their architectural characteristics.
\begin{enumerate}
\item \textit{BERT-Base:} The foundational BERT model (Bidirectional Encoder Representations from Transformers), particularly its BERT-Base version, set a new standard in natural language processing by introducing a deeply bidirectional transformer trained using two key objectives: masked language modeling (MLM) and next sentence prediction (NSP). BERT-Base is composed of 12 transformer layers, each with 12 self-attention heads, totaling 110 million parameters. This model was pre-trained on a combination of large text corpora, including BookCorpus and English Wikipedia. Its architecture and training objectives laid the groundwork for several derivative models, each designed to enhance certain aspects of BERT's performance, efficiency, or scalability.
\item \textit{RoBERTa:} RoBERTa (Robustly Optimized BERT Approach) improves upon BERT by adopting a more refined training strategy. It discards the next sentence prediction objective, which was found to have a limited impact on downstream task performance. Instead, it emphasizes a larger-scale training approach, utilizing a significantly expanded training corpus and employing dynamic masking, where the masking patterns in the MLM objective are altered across epochs. Furthermore, RoBERTa is trained over longer periods using larger batches and higher learning rates, resulting in a model that consistently outperforms the original BERT on various language understanding benchmarks.
\item \textit{DistilBERT:} DistilBERT is another derivative of BERT, designed with an emphasis on computational efficiency and speed. It is created using knowledge distillation, where a smaller student model is trained to mimic the behavior of a larger teacher model, in this case, the original BERT. By employing a combination of loss functions—distillation loss, cosine embedding loss, and language modeling loss—DistilBERT effectively captures the linguistic knowledge of BERT while being 40\% smaller and 60\% faster. Despite this compression, it retains approximately 97\% of BERT’s performance on major NLP benchmarks, making it a viable option for real-time applications.
\item \textit{ALBERT:} ALBERT (A Lite BERT) represents a distinct approach, focusing on model efficiency. It reduces the overall parameter count without a substantial loss in performance. This is achieved through two primary techniques: parameter sharing across transformer layers and factorized embedding parameterization, which separates the vocabulary embedding size from the hidden layer dimensions. Additionally, ALBERT replaces the NSP objective with a sentence order prediction (SOP) task, which has proven more effective for capturing inter-sentence coherence. These optimizations make ALBERT a lightweight model that delivers strong performance with significantly reduced computational requirements.
\end{enumerate}
To adapt BERT for question-answering, a QA output head is added on top of the pre-trained BERT model that predicts the start and end positions of the answer span within the context text. The input consists of a question and a context, formatted as \textit{[CLS] Question [SEP] Context [SEP]}. BERT processes this input to generate token representations, which the QA head then uses to predict two sets of scores—one for the start and another for the end of the answer span. During training, the model learns to accurately select the correct answer span within the context, enabling it to perform QA effectively.

\subsection{Performance Metrics}
The study adopts two distinct evaluation metrics to evaluate the performance of the targeted models. The selected evaluation metrics are designed to capture both the semantic alignment and structural similarity between the predicted and the correct output. Specifically, cosine similarity and the $F_1$ score serve as the primary metrics for assessment. \par
\subsubsection{Cosine Similarity:}

Cosine similarity serves as a measure of similarity between two vectors by calculating the cosine of the angle $(\theta)$ between them. Mathematically, cosine similarity is given as follows:

\[
\cos \theta =
\frac{\vec{a} \cdot \vec{b}}
{\|\vec{a}\| \cdot \|\vec{b}\|}
\]
\noindent

where $\vec{a}$ and $\vec{b}$ are vectors representing the embeddings of the correct answer in the dataset and the predicted answer by the model, respectively. \par
\noindent

Transformer-based models generate contextual embeddings in which semantic information is primarily represented through the orientation of embedding vectors rather than their magnitude. Consequently, cosine similarity provides an effective and computationally efficient mechanism for evaluating semantic alignment between generated and reference answers, while remaining largely scale-invariant to variations in embedding magnitude. Unlike traditional lexical overlap-based metrics such as BLEU or ROUGE, cosine similarity evaluates semantic consistency within dense contextual embedding spaces and therefore remains effective even when lexical or syntactic variations are present. This property is particularly important in the proposed reliability assessment framework, where semantically equivalent responses generated under stochastic and paraphrased perturbations may differ in wording or sentence structure while still preserving the intended meaning. When the cosine value $(\cos(\theta))$ approaches 1, it indicates strong semantic alignment between the generated and target answers, whereas values approaching 0 reflect substantial semantic divergence between the corresponding contextual representations.

\subsubsection{$F_1$ Score:} $F_1$ Score is a metric used to assess the model’s performance by balancing precision and recall. The $F_1$ Score is the harmonic mean of precision and recall, providing a balanced measure of the model's ability to make accurate predictions while also identifying as many relevant answers as possible. It is calculated as:
$$F_1=\frac{2 \times \text { Precision } \times \text { Recall }}{\text { Precision }+ \text { Recall }} $$

\subsection{Reliability Assessment}
The detailed methodology for assessing the reliability of the targeted models is described by Algorithms \ref{alg1} and \ref{alg2}. The methods assess model reliability from different perspectives. Together, they provide a comprehensive measure of the model’s stability.   \par
To evaluate the reliability of a model under internal configuration changes, we use dropout, as shown in Algorithm \ref{alg1}. The process begins by accepting several essential inputs: a pre-trained model, its corresponding tokenizer (if applicable), a dataset on which the evaluation will be conducted, a set of evaluation metrics for performance measurement, and a specified number $N$, representing the number of stochastic samples to be drawn per input instance. Following this, dropout is introduced into the model’s predictions. For each input instance from the dataset, the algorithm generates $N$ different outputs by passing the same input through the model $N$ times, each time under slightly different internal configurations due to the induced stochastic behavior.
\begin{algorithm}[htbp]
\caption{Methodology for Evaluating Model Reliability Under Internal Configuration Changes, i.e, Dropout. }
\label{alg1}
\begin{algorithmic}[1]
\State \textbf{Input:} Pre-trained model, tokenizer, dataset, evaluation metrics, number of stochastic samples ($N$)
\State \textbf{Output:} Aggregated performance scores and reliability assessment
\State Enable stochastic behavior in model predictions (e.g., Monte Carlo dropout or alternative methods).
\State Generate $N$ stochastic samples for a given input.
\For{each stochastic sample}
\State Compute the evaluation metrics with respect to the ground truth.
\EndFor
\State Compute the average value and variance of the selected evaluation metrics across all stochastic samples.
\State Aggregate the averaged scores and the variances to obtain a representative performance measure per input.
\State Repeat the process for multiple inputs and analyze the distribution of averaged scores and the variances of the performance metrics.
\State Assess the reliability of the model’s predictions under internal configuration changes.
\end{algorithmic}
\end{algorithm}
\noindent
These multiple predictions for a single input reflect the model’s variability under uncertainty. Following this, each of the $N$ predictions is evaluated against the ground truth using the predefined metrics, which capture how the model’s performance fluctuates under internal changes. After gathering the evaluation scores from all stochastic samples, the algorithm computes the average (mean) and variance of these metrics to obtain a single representative score for each input. This process is repeated for the entire dataset's input, and the resulting distribution of average scores and variances is stored.  The overall average of these values provides a summary measure of the model’s typical behavior and predictive consistency. The final assessment provides insights into the model’s reliability when subjected to internal configuration changes, ensuring a comprehensive evaluation of its stability. \par
To assess the reliability of the targeted model when handling variations in the input, a structured methodology is employed, as outlined in Algorithm \ref{alg2}. The central objective is to determine how sensitive a pre-trained model is to minor, semantically preserving changes in its inputs—such as paraphrasing or noise addition. The process begins by taking as input a pre-trained model, a dataset, a set of evaluation metrics, and a predefined threshold range that governs acceptable levels of similarity between original and modified inputs. To initiate the procedure, an individual input instance is selected from the dataset. A modified version of this input is then generated using transformation techniques, including paraphrasing a sentence, reordering phrases, or introducing controlled noise. A similarity score is computed between the original and modified input. 
\begin{algorithm}[htbp]
\caption{Methodology for Evaluating Model Reliability Under Input Variations}
\label{alg2}
\begin{algorithmic}[1]
\State \textbf{Input:} Pre-trained model, dataset, evaluation metrics, threshold range for input similarity
\State \textbf{Output:} Aggregated evaluation metric scores for reliability assessment.
\State Select an input instance from the dataset.
\State Generate a modified version of the input by applying transformations (e.g., paraphrasing, noise addition).
\State Compute the similarity between the original and modified input.
\If{similarity score falls within the predefined threshold range}
    \State Generate the model's output for the modified input.
    \State Compute the evaluation metrics between the output for the modified input and the correct output.
    \State Store the computed scores.
\EndIf
\State Analyze the collected scores to assess the model’s consistency and reliability under input variations.
\end{algorithmic}
\end{algorithm}
\noindent
The algorithm proceeds only if this score falls within a defined threshold, ensuring that the change is both significant and semantically consistent. The modified input is fed into the model to generate a new output, which is then compared against the correct output using predefined evaluation metrics. These metrics reflect how consistently the model responds to semantically similar inputs, serving as an indicator of its stability. Scores are collected across the dataset, and their averages provide an overall measure of the model’s consistency to input variations across the entire dataset, enabling a structured assessment of its reliability when subjected to handling input variations.

\subsection{Experimental Settings}
The experimental setup for this study was implemented in a Python environment. The two proposed methodologies involve multiple hyperparameters, including evaluation metrics, the number of stochastic samples, the threshold range, the pre-trained model, the tokenizer, and the dataset. While the dataset and pre-trained models used in this study have already been specified, the choice of hyperparameters is done as follows. \par
For the first methodology, the number of stochastic samples, $N$, was set to 50. This selection was based on empirical observations from experiments conducted with varying numbers of stochastic samples, ranging from 10 to 100 in increments of ten, which demonstrated that there is no variation in the model’s performance on increasing the number of samples, as it consistently produced semantically similar responses across multiple stochastic runs. Given this stability, we empirically selected a mid-range value of 50 that ensured an optimal trade-off for computational feasibility because a higher number of stochastic samples introduced significant computational cost, while a lower number could lead to potential loss of information when averaging the evaluation metrics. \par
To incorporate stochastic behavior, Monte Carlo (MC) dropout was activated during the prediction phase. Dropout \cite{24} is a widely recognized regularization technique in which a fraction of neurons is randomly dropped off during training, compelling the network to develop more robust and generalized features rather than relying on specific neurons. Instead of training, dropout was also applied during the testing phase, referred to as Monte Carlo dropout, enabling stochastic sampling from the model’s output distribution. This enabled an assessment of the model’s reliability under variations in its internal configuration.         \par
To implement the methodology for input variation, paraphrasing has been selected. Paraphrasing can be achieved through various approaches \cite{17, 18, 19}, including synonym substitution and structural modifications. Additionally, several models have been developed to facilitate the process of paraphrasing. In our methodology, we employ a pre-trained BART-based paraphrasing model (eugenesiow/bart-paraphrase) to generate semantically equivalent rephrasings using \textit{text2text-generation} pipeline from the Hugging Face Transformers library. Additionally, the Sentence-BERT model (S-BERT), all-MiniLM-L6-v2, was loaded to compute the sentence embeddings. \par
For each input question, a paraphrased version is generated and its semantic similarity with the original question is computed using cosine similarity between S-BERT embeddings. To preserve semantic consistency while still introducing meaningful linguistic variation, a similarity interval of 0.75–0.98 was adopted during paraphrase filtering. These values should not be interpreted as uniquely optimal thresholds, but rather as practically motivated operational boundaries for controlling semantic preservation and perturbation diversity. Similarity values substantially below 0.75 increasingly exhibited semantic deviation, whereas the upper bound was selected to ensure that the generated paraphrases still undergo meaningful linguistic change. Consequently, similarity values approaching 1.0 frequently produced near-duplicate paraphrases with minimal lexical variation. The validity of the proposed similarity interval was empirically assessed through a 10\% random sample of answerable questions drawn from the SQuAD dataset. The analysis examined semantic alignment using cosine similarity and lexical overlap using Jaccard similarity across different similarity regions. The findings indicated that paraphrases within the selected interval maintained strong semantic consistency while still introducing moderate lexical variation, whereas lower similarity ranges exhibited greater semantic deviation, and similarity values above 0.98 corresponded to trivial reformulations with excessively high lexical overlap. A structured overview of the similarity characteristics across distinct similarity regions is provided in Table~\ref{tab:similarity_validation}. \par
\begin{table}[ht]
\centering
\caption{ Empirical validation of the selected semantic similarity interval using cosine similarity and Jaccard similarity analysis.}
\label{tab:similarity_validation}
\renewcommand{\arraystretch}{1.3}
\resizebox{\textwidth}{!}{%
\begin{tabular}{cccp{5.5cm}}
\hline
\textbf{Similarity Range} & \textbf{Avg Cosine Similarity} & \textbf{Avg Jaccard Similarity} & \textbf{Interpretation} \\ \hline
$<$0.75 & 0.7078 & 0.2932 & Higher semantic deviation \\ 
0.75--0.98 & 0.9206 & 0.5642 & Balanced semantic preservation and lexical variation \\ 
$>$0.98 & 0.9971 & 0.9387 & Near-duplicate reformulations \\ \hline
\end{tabular}%
}
\end{table}
Following the similarity-based filtering process, if a generated paraphrased question satisfies the specified threshold interval of 0.75 – 0.98, the model predicts answers for the paraphrased questions, referred to as paraphrased answers, and the results are compared against the correct answers in the dataset by computing the cosine similarity and $F_1$ score between the correct and paraphrased answers, which provides a measure of the model’s sensitivity towards input variation. The resulting distribution of cosine similarity and $F_1$ scores across the entire dataset is analyzed to assess the reliability of the targeted models. The average values from these distributions indicate the degree of alignment between the paraphrased answers and the ground-truth responses for the entire dataset. An average value approaching 1 suggests that the majority of paraphrased answers closely match the correct answers, thereby reflecting the model’s stability in handling variations in input phrasing and its ability to maintain semantic consistency despite changes in the representation of the input. To ensure reproducibility and transparency of the proposed reliability evaluation framework, the implementation code has been made publicly available through the following GitHub repository;
\url{https://github.com/uncertainity-quantification/reliability-estimation-qa1}

\section{Results}
\label{res}
This section presents a comprehensive analysis of the results obtained from our experiments. It includes findings from both methodologies evaluating model reliability. The results provide insights into how changes in model configuration and input affect the stability and consistency of the model’s predictions, ultimately contributing to a deeper understanding of the model’s reliability in question-answering tasks.

\subsection{Effect of Dropout Rate Variation on Answer Consistency}

As Monte Carlo Dropout is employed to introduce stochasticity in the model, it inherently involves a key hyperparameter - the selection of an appropriate dropout rate. Before incorporating dropout during the prediction phase to assess the model's reliability, our initial objective is to determine an appropriate dropout rate that can be effectively utilized. To achieve this, we experimented with multiple dropout rates, systematically varying the percentage to observe its influence on the model's predictions. \par
To determine an appropriate dropout rate, the methodology outlined in Algorithm~1 is applied across a range of dropout rates, varying from 0\% (no dropout at all) to a maximum of 35\% dropout during the prediction phase for all models. The average results for all models with various dropout rates across both datasets are presented in Figure~3 and Figure~4, showing their impact on the model's predicted outputs through semantic similarity scores. The findings highlight how increasing dropout influences the consistency of the predicted answers. As dropout increases, the semantic similarity between the predicted and ground-truth answers generally declines for the answerable questions of both datasets, indicating greater variability and uncertainty in the model's responses. However, for unanswerable questions, an increasing trend is observed, as higher dropout rates lead to more blank predictions, which are technically correct. This results in an artificial rise in performance for unanswerable cases and a concurrent decline in semantic alignment for answerable ones. \par

\begin{figure}[H]
    \centering
    \includegraphics[width=\textwidth]{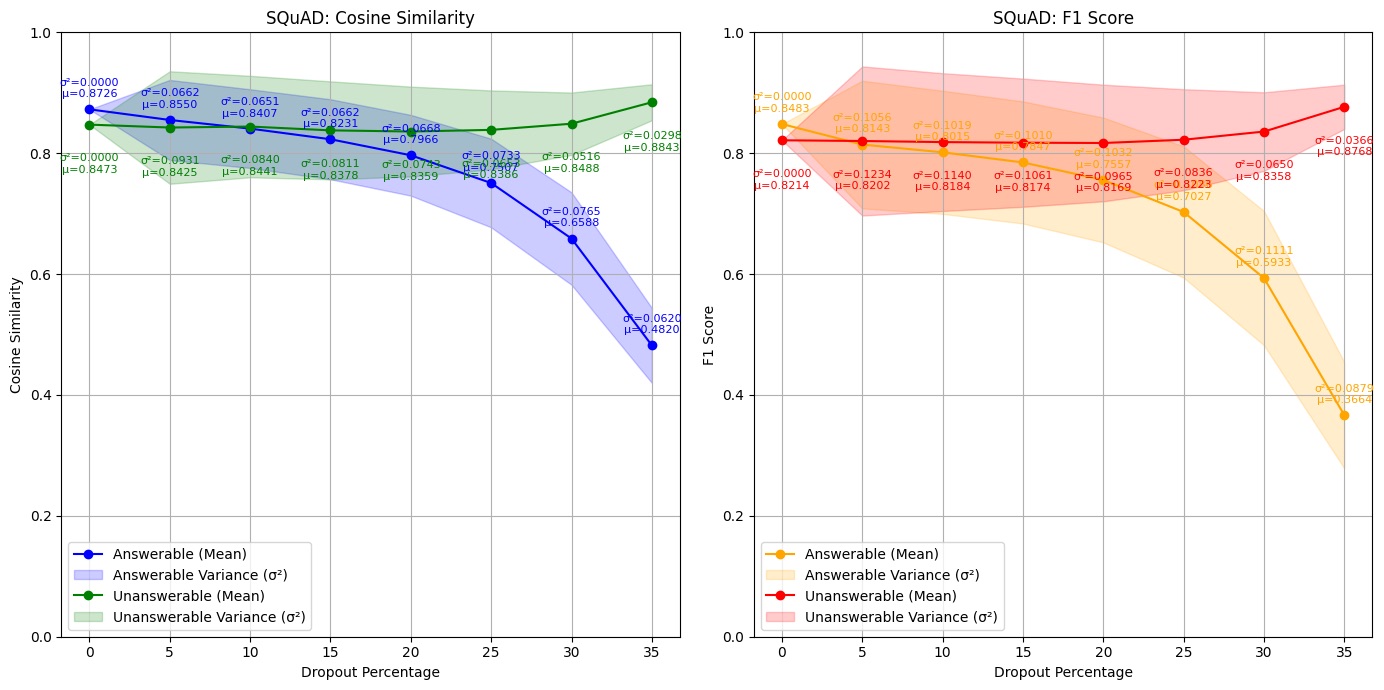}
    \caption{Analyzing Output Stability of SQuAD Under Varying Dropout Percentages.}
    \label{jkl}
  \end{figure}
  
  \begin{figure}[H]
    \centering
    \includegraphics[width=\textwidth]{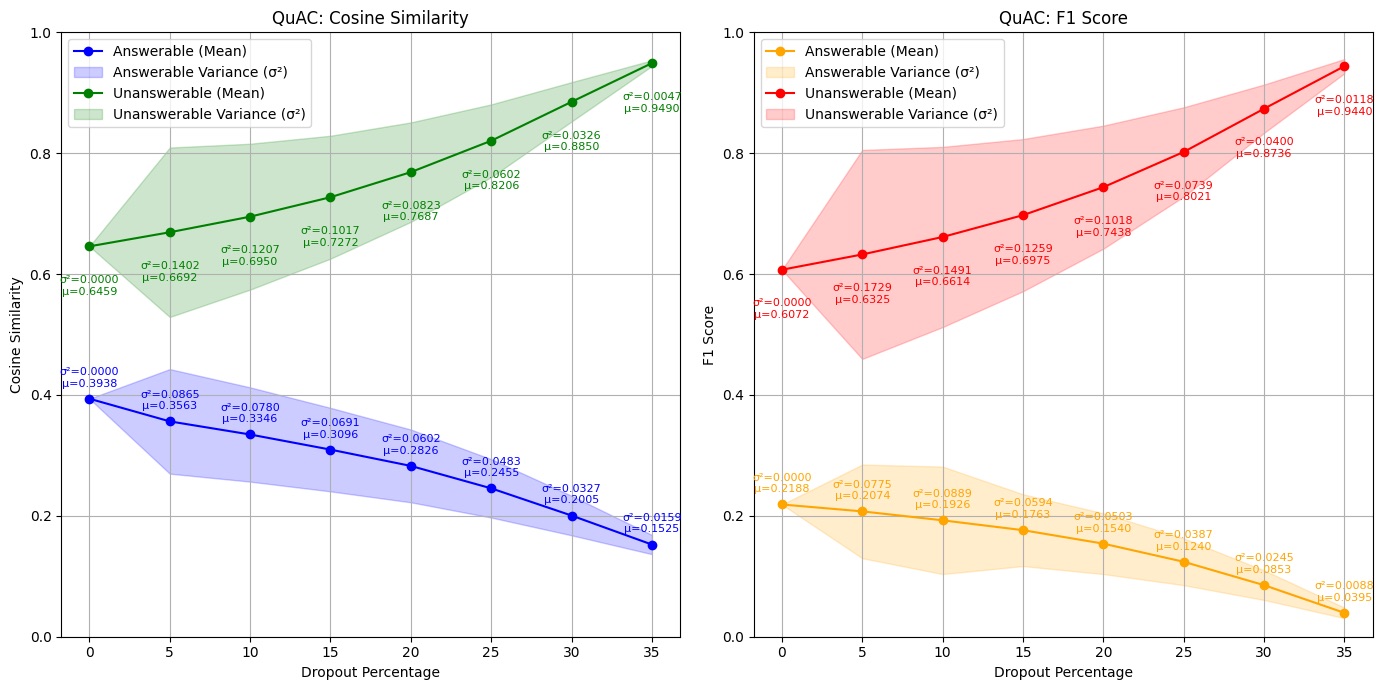}
    \caption{Analyzing Output Stability of QuAC Under Varying Dropout Percentages.}
    \label{xcv}
\end{figure}

 The analysis also revealed a notable shift in the model's behavior from a dropout rate of 15\% and beyond, particularly in its handling of answerable and unanswerable questions across both datasets as the models increasingly produced blank responses, indicating information loss. To avoid this imbalance, dropout rates $\geq 15\%$ are excluded. Moreover, the results show that the mean values of both evaluation metrics remain nearly unchanged between 0\% and 5\% dropout, indicating that dropping 5\% of neurons in the models does not introduce the desirable stochastic variations in the model's internal dynamics. \par
As further evidenced by the variance bands illustrated in Figures~3 and~4, the spread of performance scores at 5\% dropout remains notably narrow and closely mirrors the distribution observed under the 0\% dropout condition, corroborating the insufficiency of this rate in inducing meaningful internal stochastic diversity. In contrast, the variance band at 10\% dropout exhibits a discernible widening across both datasets, indicating that this rate introduces sufficient variability in model behaviour without causing substantial degradation in semantic alignment. This empirical observation is consistent with \cite{7}, where a comparable dropout rate was employed within a Monte Carlo Dropout-based reliability estimation framework, although without an explicit empirical justification for its selection. The present study addresses this gap by offering a systematic, data-driven rationale derived from a comprehensive sweep of dropout rates across two diverse QA datasets, thereby establishing a more principled basis for the adoption of this hyperparameter. Therefore, a 10\% dropout rate is selected as a balanced operating point that introduces sufficient stochastic variability while maintaining stable semantic consistency in the generated responses.

\subsection{Evaluating Reliability Under Dropout-Induced Variation}
By selecting an appropriate dropout rate, we assess the reliability of all targeted models by enabling dropout in the prediction phase. The reliability of the models, measured through both cosine similarity and $F_1$ score standard deviation, 
reveals crucial differences in their consistency across the SQuAD and QuAC datasets. Table \ref{tab:SQuAD} and \ref{tab:QuAC} presents the evaluation results of four BERT model variants on two datasets, SQuAD and QuAC. Since both datasets include answerable and unanswerable questions, each targeted model was evaluated separately for these categories. The results for answerable questions are shown in the row labeled \textit{Ans}, while those for unanswerable questions are presented in the row labeled \textit{No-Ans}. The combined performance of the models for both categories within each dataset is summarized in the row labeled \textit{Total}. The table reports the mean values of cosine similarity and $F_1$ scores for each model, along with the average standard deviations of cosine similarity and $F_1$ scores, which reflect the reliability of these models under Monte Carlo dropout. The model's accuracy was assessed by comparing the predicted answers from the unperturbed models to the ground-truth responses provided in the dataset. To further examine the impact of internal stochasticity, accuracy was also evaluated with Monte Carlo Dropout enabled during inference—referred to as "MCD Accuracy"—to capture performance variations between the perturbed and unperturbed models. Instead of relying on exact textual matches, a prediction was classified as an exact match if the semantic similarity score between the predicted answer and the ground truth was greater than or equal to 0.95. Using this criterion, the accuracy of all four unperturbed and perturbed models was assessed across both datasets.   \par
The comparative evaluation presented in Tables \ref{tab:SQuAD} and \ref{tab:QuAC} offers a comprehensive analysis of the performance and consistency of the four transformer-based models. Notably, the accuracy of the perturbed model and the unperturbed model is comparable, suggesting that stochastic inference does not degrade overall performance, which highlights the consistency of BERT-based models. To assess reliability, average values of cosine similarity and $F_1$ scores were analyzed to determine the semantic alignment between predicted and ground truth answers. In parallel, the standard deviation associated with these metrics provided insight into output variability introduced by internal stochastic perturbations, quantifying model consistency. \par
On the SQuAD dataset, DistilBERT demonstrated superior semantic coherence for answerable questions, achieving the highest average cosine similarity (0.8992) and $F_1$ score (0.8496) with moderate standard deviations (0.0906 and 0.1149, respectively), suggesting stable and accurate behavior. 
\begin{table}[H]
\centering
\caption{Comparative Results of Model Accuracy and Consistency on SQuAD Dataset. 
}
\label{tab:SQuAD}
\resizebox{\textwidth}{!}{
\begin{tabular}{cccccc}
\hline
Models &Category & Acc.& MCD Acc. & Cosine Similarity & $F_1$ Score \\
&&(\%) &(\%) &(avg $\pm$ std)&(avg $\pm$ std)\\
\hline
&Ans  & 75.29 & 74.78 & 0.856 $\pm$ 0.1404  & 0.8006 $\pm$ 0.1783 \\
RoBERTa &No-Ans & \textbf{81.85} & \textbf{81.88} & \textbf{0.8407 $\pm$ 0.167} & \textbf{0.8190 $\pm$ 0.1778} \\
&Total & \textbf{78.57} & \textbf{78.34} & \textbf{0.8481 $\pm$ 0.1543} & \textbf{0.8098 $\pm$ 0.1772} \\
\\
 &Ans & 59.85 & 59.60 & 0.7471 $\pm$ 0.1833 & 0.6135 $\pm$ 0.2166 \\
BERT-Base &No-Ans & 77.44 & 76.87 & 0.7915 $\pm$ 0.2142 & 0.7676 $\pm$ 0.2236 \\
 &Total & 68.66 & 68.25 & 0.7605 $\pm$ 0.1992 & 0.6907 $\pm$ 0.2200 \\
\\
&Ans& \textbf{77.75} & \textbf{77.82} & \textbf{0.8992 $\pm$ 0.0906} & \textbf{0.8496 $\pm$ 0.1149} \\
DistilBERT &No-Ans & 79.70 & 79.57 & 0.8184 $\pm$ 0.1319 & 0.8039 $\pm$ 0.1407 \\
&Total & 78.21 &  78.19 & 0.7912 $\pm$ 0.1131 & 0.7584 $\pm$ 0.1285 \\
\\
&Ans & 56.92 & 56.36 & 0.7403 $\pm$ 0.2254 & 0.6182 $\pm$ 0.2687 \\
AlBERT &No-Ans & 78.81 & 78.70 & 0.8089 $\pm$ 0.2449 & 0.7893 $\pm$ 0.2565 \\
&Total &  67.88 & 67.55 & 0.7746 $\pm$ 0.2354 & 0.7038 $\pm$ 0.2627 \\
\hline
\end{tabular}
}
\end{table}
\noindent
RoBERTa, meanwhile, showed strong performance in handling unanswerable inputs with a cosine similarity of 0.8407 $\pm$ 0.167, and $F_1$ score of 0.8190 $\pm$ 0.1778, dominating in overall consistency, with average scores of 0.8481 (cosine similarity) and 0.8098 ($F_1$), their respective standard deviations being 0.1543 and 0.1772, reflecting its balanced handling across both input categories.  \par
In contrast, the QuAC dataset revealed a different pattern. While semantic accuracy for answerable questions was generally lower across all models, RoBERTa achieves the best semantic alignment (cosine similarity 0.3641 $\pm$ 0.1546, $F_1$ score 0.1942 $\pm$ 0.1414) in this category. For unanswerable queries, AlBERT and DistilBERT show strong performance, both exceeding a cosine similarity score of 0.81. However, ALBERT exhibited substantially higher standard deviation (0.2820) than DistilBERT (0.1775), reflecting less stable performance. Considering overall metrics, RoBERTa and DistilBERT displayed strong consistency, with comparable values of cosine similarity (0.4307 and 0.4265) and $F_1$ score (0.2830 and 0.2881), respectively. However, there is a significant difference captured in their standard deviations (0.1761, 0.1761, and 0.1470, 0.1360), with DistilBERT showing less dispersion.
\begin{table}[H]
\centering
\caption{Comparative Results of Model Accuracy and Consistency on QuAC Dataset.
}
\label{tab:QuAC}
\resizebox{\textwidth}{!}{
\begin{tabular}{cccccc}
\hline
Models &Category & Acc. & MCD Acc. & Cosine Similarity & $F_1$ Score \\
& &(\%) & (\%) &(avg $\pm$ std)&(avg $\pm$ std)\\
\hline  
&Ans & \textbf{10.17} &  \textbf{10.24} & \textbf{0.3641 $\pm$ 0.1546} &\textbf{ 0.1942 $\pm$ 0.1414} \\
RoBERTa & No-Ans & 65.68 & 65.95  & 0.6938 $\pm$ 0.2431 & 0.6591 $\pm$ 0.2650 \\
& Total & 21.39 & 21.50 & \textbf{0.4307 $\pm$ 0.1761} & 0.2830 $\pm$ 0.1761 \\
\\
& Ans & 6.37 & 6.24 & 0.3087 $\pm$ 0.1718 & 0.1376 $\pm$ 0.1456 \\
BERT-Base & No-Ans & 71.06 & 73.62 & 0.7528 $\pm$ 0.2642 & 0.7256 $\pm$ 0.2891 \\
& Total & 19.45 & 19.85 & 0.3984 $\pm$ 0.1942 & 0.2564 $\pm$ 0.1838 \\
\\
&  Ans & 8.03 & 8.03 & 0.3224 $\pm$ 0.1382 & 0.1482 $\pm$ 0.1149 \\
DistilBERT & No-Ans & 81.76 & \textbf{82.84} & \textbf{0.8378 $\pm$ 0.1775} & 0.8156 $\pm$ 0.1786 \\
& Total & \textbf{22.93} & \textbf{23.14} & 0.4265 $\pm$ 0.1470 & \textbf{0.2881 $\pm$ 0.1360} \\
\\
& Ans & 0.05 & 0.07 & 0.3105 $\pm$ 0.2012 & 0.1110 $\pm$ 0.1261 \\
AlBERT & No-Ans & \textbf{81.97} & 81.63 & 0.8309 $\pm$ 0.2615 & \textbf{0.8161 $\pm$ 0.2820} \\
 & Total & 16.60 & 16.55 & 0.4156 $\pm$ 0.2147 & 0.2535 $\pm$ 0.1697 \\
\hline
\end{tabular}
}
\end{table}
\noindent
These results collectively indicate that RoBERTa and DistilBERT consistently provide high semantic alignment with comparable low standard deviations across both datasets, making them appropriate candidates for tasks that prioritize reliability. In contrast, BERT-Base demonstrates moderate and balanced results but demonstrates comparatively lower performance in overall consistency. ALBERT, though strong on unanswerable questions, performs poorly on answerable ones, especially on the QuAC dataset, where it records an answerable accuracy as low as 0.07\%, making it less suitable for nuanced tasks. Therefore, model selection should be guided not only by average accuracy but also by consistency metrics that capture behavior under input and configuration variation.
\par
\begin{figure}[H]
  \centering
  \begin{minipage}[b]{0.45\textwidth}
    \includegraphics[width=\textwidth]{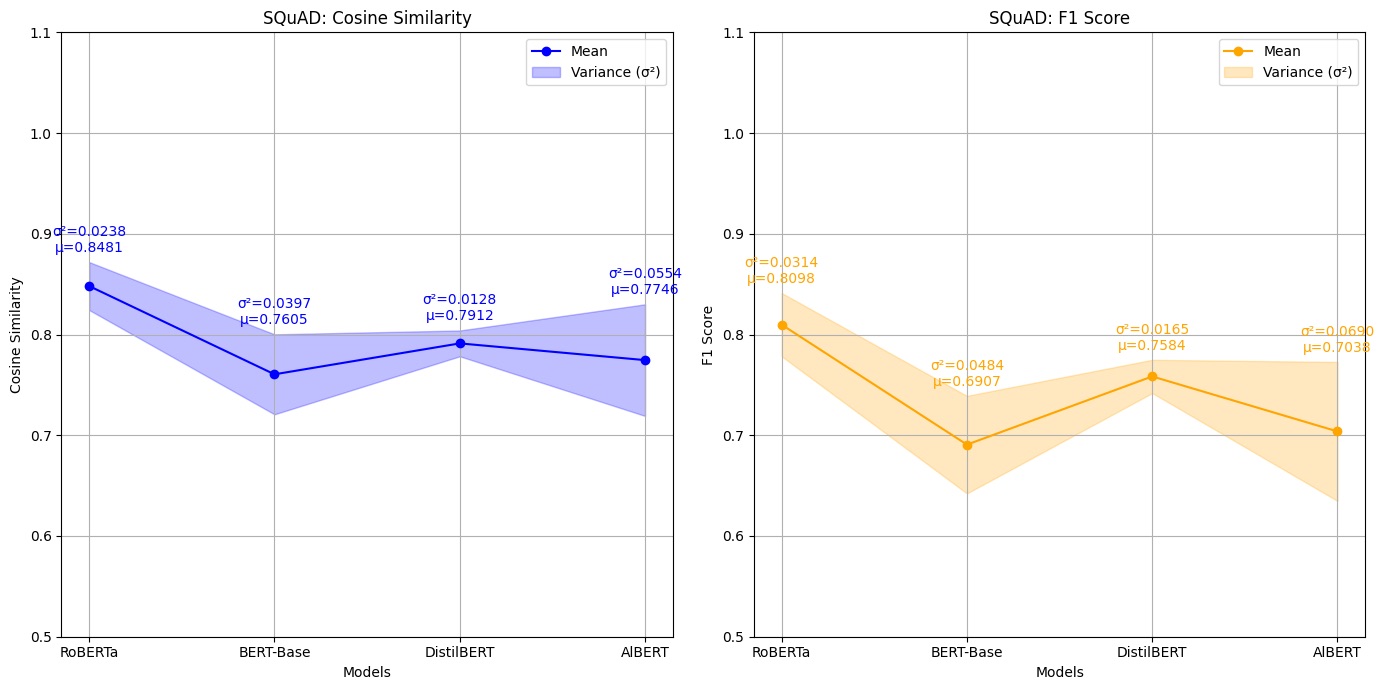}
 \caption{Comparison of Cosine Similarity and $F_1$ Scores for SQuAD Across BERT Models.}
    \label{gas}
  \end{minipage}
  \hfill
  \begin{minipage}[b]{0.45\textwidth}
    \includegraphics[width=\textwidth]{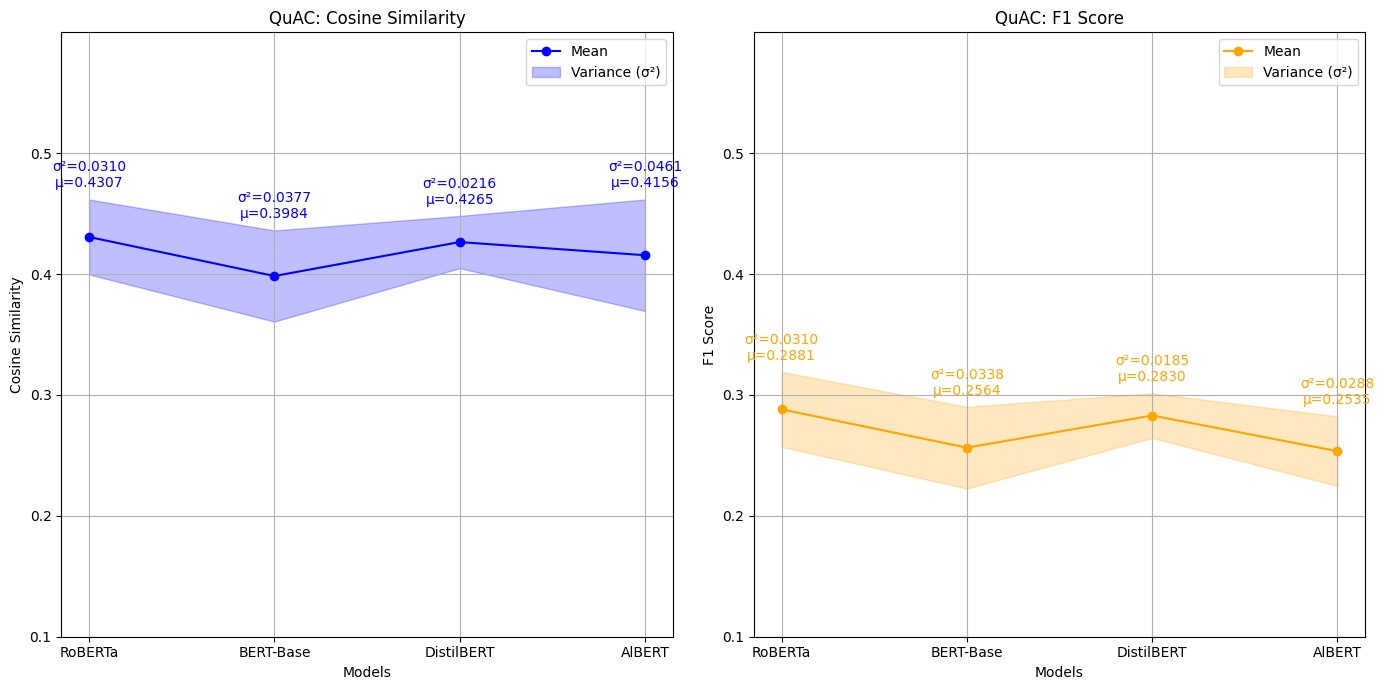}
    \caption{Comparison of Cosine Similarity and $F_1$ Scores for QuAC Across BERT Models.}
    \label{gap}
  \end{minipage}
\end{figure}

Figures \ref{gas} and \ref{gap} illustrate the pictorial representation of the results of Tables \ref{tab:SQuAD} and \ref{tab:QuAC}, showing the overall performance of the models across the SQuAD and QuAC datasets. The visual trends suggest that RoBERTa and DistilBERT maintain higher evaluation metric scores, indicating that the predicted stochastic outputs are closely aligned with the correct ones from the dataset, thereby reflecting consistent predictive performance. Moreover, the narrow spread in their outputs, as seen in the variance bands, highlights their low variability, which is indicative of strong consistency and reliability. In contrast, AlBERT, though moderately reliable, exhibits fluctuations, especially in unanswerable cases, making it less stable. While BERT-Base lags significantly, reflecting its limitations in both reliability and accuracy. Furthermore, the comparatively similar performance trends observed between RoBERTa and DistilBERT raise the question of whether the observed differences are statistically significant or due to random variation. 
\noindent

To investigate this, Welch’s t-test is employed at a significance level of 5\%, as it is well-suited for comparing the means of two independent samples that may have unequal variances. The statistical hypotheses for this test are formulated as follows: \\  \vspace{0.1cm}
\textit{Null Hypothesis ($H_0$): There is no significant difference in the performance of RoBERTa and DistilBERT under internal configuration changes.} \vspace{0.1cm} \\ 
\textit{Alternative Hypothesis ($H_1$): There is a significant difference in the performance of RoBERTa and DistilBERT under internal configuration changes.} \\   \vspace{0.1cm}
\noindent
For the SQuAD dataset, the test yields a p-value of 0.0 ($<$ 0.05), leading to the rejection of the null hypothesis, confirming that there is a statistically significant difference in the performance of RoBERTa and DistilBERT under internal stochastic variations. In contrast, on the QuAC dataset, the p-value obtained is 0.4547, which exceeds the 0.05 threshold. Consequently, we fail to reject the null hypothesis, suggesting that both models perform similarly when exposed to internal perturbations. These outcomes indicate that the performance distinction between models is context-dependent and reinforces the importance of evaluating model robustness in a dataset-specific manner.

\par
A key concern in evaluating model reliability is understanding whether enabling Monte Carlo dropout during inference significantly changes the model’s structure. If enabling dropout leads to substantial deviations in the generated responses, it could suggest a significant disruption to the learned representations, thereby raising concerns regarding the stability and trustworthiness of the model’s predictions. To systematically investigate this effect, we compute the correlation coefficients between the responses produced by the unperturbed model and those obtained after activating dropout, as detailed in Table \ref{tab:semantic-metric}. To compute accuracy, the unperturbed model’s predictions are evaluated against correct answers. For the perturbed model, evaluation metrics are averaged across stochastic samples to obtain a single score per input. This is done for the entire dataset, and the correlation between the resulting scores is analyzed to evaluate the effect of perturbations on model predictions. These coefficients provide a quantitative measure of output consistency, where values approaching 1 signify strong agreement between perturbed and unperturbed predictions, reinforcing the stability of the underlying model configuration.  \par
\begin{table}[H]
\centering
\caption{Semantic metric comparison across models on SQuAD and QuAC datasets.}

\begin{tabular}{|c|c|c|c|}   \hline
\diagbox{Models}{Datasets} & Semantic Metric     & SQuAD & QuAC \\ \hline
\multirow{2}{*}{RoBERTa}    & Cosine Similarity   & 0.8573 & 0.8607 \\ \cline{2-4} 
                            & $F_1$ Score           & 0.8673 & 0.8784 \\ \hline
\multirow{2}{*}{BERT-Base}  & Cosine Similarity   & 0.9036 & 0.8941  \\ \cline{2-4} 
                            & $F_1$ Score           & 0.9177 & 0.9071 \\ \hline
\multirow{2}{*}{DistilBERT} & Cosine Similarity   & \textbf{0.9464 }& 0.9186 \\ \cline{2-4} 
                            & $F_1$ Score           & 0.9423 & \textbf{0.9343} \\ \hline
\multirow{2}{*}{AlBERT}     & Cosine Similarity   & 0.9054  & \textbf{0.9316}  \\ \cline{2-4} 
                            & $F_1$ Score           &\textbf{ 0.9453 } & 0.9026  \\ \hline
\end{tabular}
\label{tab:semantic-metric}
\end{table}
The empirical results consistently show high correlation coefficients across all models and datasets, indicating that dropout has a minimal impact on prediction stability.  RoBERTa maintains strong correlations on SQuAD (0.8573 cosine similarity, 0.8673 $F_1$) and QuAC (0.8607, 0.8784), while BERT-Base exhibits even higher values, exceeding 0.90 in most cases, while DistilBERT and ALBERT also demonstrate stability. These findings highlight the resilience of learned representations under stochastic perturbations. \par
These results underscore the non-disruptive nature of Monte Carlo dropout, affirming that its integration does not substantially alter the model’s inference. The high consistency between the unperturbed and perturbed predictions substantiates the viability of dropout-based uncertainty estimation as a reliable framework for evaluating prediction stability under stochastic perturbations.. Moreover, the stability of correlation values across datasets—despite their accuracy differences on SQuAD and QuAC—suggests that dropout-driven reliability assessments generalize effectively across diverse QA paradigms. This reinforces the broader applicability of Monte Carlo dropout as a principled approach to quantifying model uncertainty, offering insights into predictive stability while maintaining the integrity of learned representations.

\subsection{Evaluating Reliability Under Input-Induced Variation}
The study employs the proposed methodological framework systematically to introduce controlled input perturbations. Paraphrasing is adopted to introduce controlled perturbations in the input questions using a pre-trained paraphrasing model. The algorithm calculates the evaluation metric (cosine similarity, $F_1$ score) for the paraphrased answer against the correct answer provided in the dataset. This score quantifies the similarity between the paraphrased and correct answers, where a higher value indicates greater reliability, while a lower value suggests a deviation in the answer. This process is applied to the entire dataset, including answerable and unanswerable questions, and the average scores of evaluation metrics are recorded. Table \ref{asd} presents the paraphrasing results, reporting the average values of cosine similarity and the $F_1$ scores across both datasets.   
\par
On the SQuAD dataset, RoBERTa demonstrates the strongest overall performance, achieving the highest average values of both cosine similarity and $F_1$ score across all evaluation categories. Specifically, it attains a cosine similarity of 0.829 and an $F_1$ score of 0.7503 for answerable questions. Even in the No-Answer category, where performance typically drops significantly, RoBERTa records the highest cosine similarity (0.1394) and $F_1$ score (0.0005) among all models. Its overall performance (0.4895-cosine similarity, 0.3812 - $F_1$ score) further underscores its stability to input variation, followed by AlBERT in all three categories. Additionally, DistilBERT slightly outperforms BERT-Base in total scores (0.4672 vs. 0.4630 cosine similarity; 0.3556 vs. 0.3499 $F_1$ score). Both models perform reasonably well on answerable questions but show diminished effectiveness in handling unanswerable cases. In contrast, ALBERT, while trailing RoBERTa on SQuAD, shows strong performance on the QuAC dataset, achieving the highest cosine similarity (0.3311) and $F_1$ score (0.2825) for answerable questions, and also leads in overall QuAC performance, with a cosine similarity of 0.3217 and $F_1$ score of 0.2333, indicating greater consistency to paraphrasing in this dataset compared to other models. In this case as well, RoBERTa continues to deliver performance comparable to ALBERT across all three evaluation categories, achieving overall scores of 0.3206 for cosine similarity and 0.2135 for $F_1$ score. In contrast, BERT-Base and DistilBERT fall short of the performance demonstrated by both ALBERT and RoBERTa in handling paraphrased inputs on the QuAC dataset. \par
Overall, RoBERTa shows the strongest performance on the SQuAD dataset when exposed to inputs, whereas ALBERT performs best on the QuAC dataset. Both models consistently outperform BERT-Base and DistilBERT, establishing themselves as the most reliable architectures in terms of input variation. However, the performance distinction between RoBERTa and ALBERT is comparatively less pronounced. In QuAC, although RoBERTa performs marginally better in terms of evaluation metrics, ALBERT follows closely.    \par
\begin{table}[H]
\centering
\caption{Comparative Results of Cosine Similarity on Model Performance Against Paraphrased Inputs.}
\label{asd}
\begin{tabular}{cccccc}
\hline
Models & Category &  \multicolumn{2}{c}{SQuAD}& \multicolumn{2}{c}{QuAC}   \\ \hline
    &    & Cosine & $F_1$ & Cosine & $F_1$ \\
    &   & Similarity& Score & Similarity & Score \\   \hline  
&Ans &\textbf{ 0.829} & \textbf{0.7503} & 0.3278 & 0.2546 \\
RoBERTa & No-Ans  & \textbf{0.1394}& \textbf{0.0005}& \textbf{0.1510} & \textbf{0.0159}  \\
& Total & \textbf{0.4895} & \textbf{0.3812}   & 0.3206 & 0.2135  \\
\\
& Ans & 0.7848 & 0.6891 & 0.3086 &0.2286  \\
BERT-Base & No-Ans & 0.1310 & 0.0  & 0.1486 & 0.0053 \\
& Total & 0.4630 & 0.3499 & 0.3012 &  0.1897 \\
\\
&  Ans & 0.7915 &0.7033  & 0.3004 &0.2249  \\
DistilBERT & No-Ans & 0.1327 & 0.0  & 0.1331 & 0.0018 \\
& Total & 0.4672 & 0.3556  & 0.2930 & 0.1860 \\
\\
& Ans & 0.8014 & 0.6346  & \textbf{0.3311} & \textbf{0.2825} \\
AlBERT & No-Ans & 0.1323  & 0.0 & 0.1197 & 0.0 \\
 & Total & 0.4720 & 0.3222  & \textbf{0.3217} & \textbf{0.2333} \\
\hline
\end{tabular}
\end{table}
\noindent

Although the evaluation metrics demonstrate variations in performance, several model pairs exhibit closely aligned results across both datasets. These closely aligned results necessitate a formal statistical analysis to determine whether the observed differences are statistically meaningful. Therefore, pairwise Wilcoxon signed-rank tests are conducted across all model combinations. \par
\begin{figure}[H]
\centering
\begin{subfigure}[t]{0.48\textwidth}
    \centering
    \includegraphics[width=\linewidth]{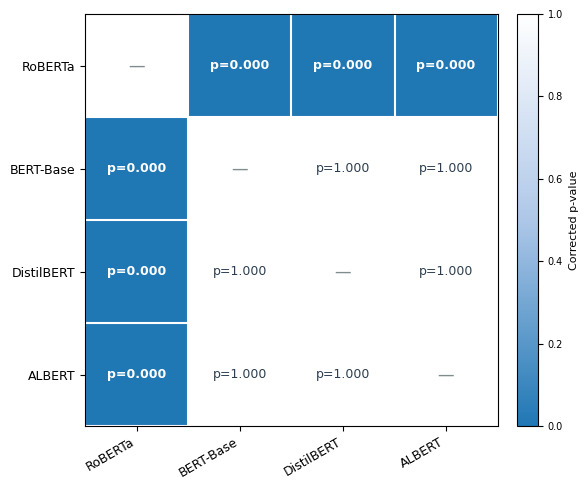}
    \caption{SQuAD dataset — Cosine Similarity}
\end{subfigure}
\hfill
\begin{subfigure}[t]{0.48\textwidth}
    \centering
    \includegraphics[width=\linewidth]{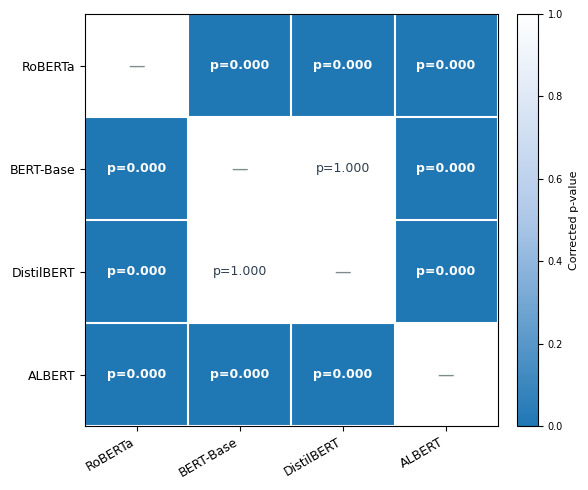}
    \caption{SQuAD dataset — F1 Score}
\end{subfigure}
\hspace{0.01\textwidth}
\begin{subfigure}[t]{0.48\textwidth}
    \centering
    \includegraphics[width=\linewidth]{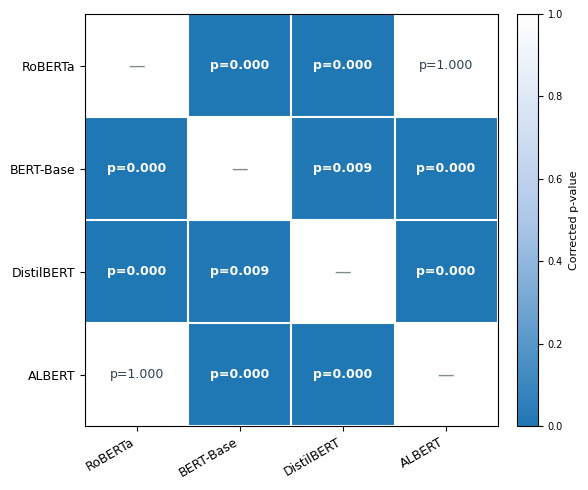}
    \caption{QuAC dataset — Cosine Similarity}
\end{subfigure}
\hfill
\begin{subfigure}[t]{0.48\textwidth}
    \centering
    \includegraphics[width=\linewidth]{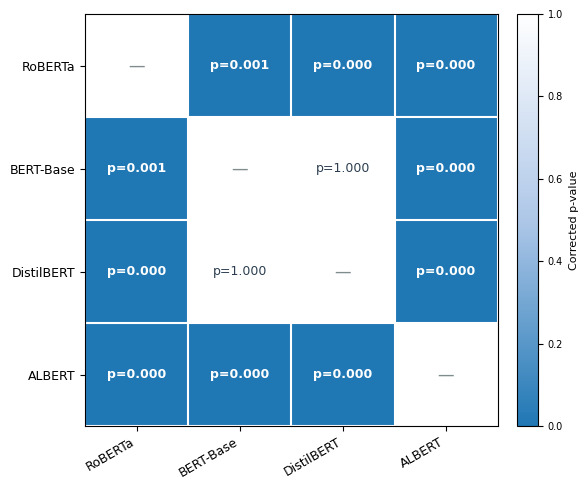}
    \caption{QuAC dataset — F1 Score}
\end{subfigure}
\captionsetup{justification=raggedright, singlelinecheck=false}
\caption{Pairwise Wilcoxon signed-rank test heatmaps with Bonferroni-corrected p-values comparing RoBERTa, BERT-Base, DistilBERT, and ALBERT across SQuAD and QuAC datasets.}
\label{fig:em_comparison}
\end{figure}

Figure ~\ref{fig:em_comparison} illustrates the pairwise Wilcoxon signed-rank analysis across all model combinations for both datasets under cosine similarity and F1 score metrics. Given four evaluated architectures, two datasets, and two evaluation metrics, the statistical framework comprises 24 pairwise hypothesis tests (6 model pairs $\times$ 2 datasets $\times$ 2 metrics). To mitigate the inflation of Type-I error induced by multiple comparisons, Bonferroni correction is employed, yielding a corrected significance threshold of $\alpha = 0.05/24 \approx 0.0021$. The heatmaps in Figure ~\ref{fig:em_comparison} depict the Bonferroni-adjusted p-values corresponding to each pairwise comparison, where darker regions denote statistically significant performance disparities. Several corrected p-values attain extremely small magnitudes and consequently appear as 0.000 following rounding to four decimal places, whereas values approaching unity are represented as 1.000. The statistical evidence indicates the presence of significant performance discrepancies across multiple model pairs under paraphrased input perturbations, while certain comparisons remain statistically indistinguishable after correction. Collectively, these findings further emphasize the dataset-specific and architecture-dependent nature of reliability under semantic input variation.
    
Human evaluation was also conducted to check the correctness of the results obtained by automatic metrics. In the proposed methodology, a response is considered correct only if it exactly matches the ground truth provided in the dataset. Consequently, there is no ambiguity when a predicted answer is marked correct, as it precisely aligns with the ground truth. The primary focus of the human annotation process, therefore, was to re-examine responses that were automatically classified as incorrect, as such predictions may still exhibit partial or complete semantic alignment with the reference answers. To facilitate this assessment, responses were categorized into three distinct classes: \textit{correct}, \textit {partially correct}, and \textit{incorrect}. A response was classified as "\textit{correct}" if it was semantically identical to the ground truth, irrespective of lexical or syntactic variations; "\textit{partially correct}" if it conveyed a portion of the intended meaning but lacked completeness; and "\textit{incorrect}" if it diverged entirely from the ground truth in both semantic and lexical aspects, containing irrelevant or inconsistent information. For this evaluation, two models with contrasting performance levels were selected: RoBERTa (high-performing and reliable) and DistilBERT (lower-performing and less reliable).  \par

To ensure a balanced and representative evaluation, a stratified random 10\% subset of the responses was selected from both the SQuAD and QuAC datasets, including equal proportions of answerable and unanswerable questions. Human annotation was independently performed by three annotators following predefined evaluation criteria. To quantify inter-annotator agreement, Fleiss’ Kappa coefficient $(\kappa)$ is employed \cite{landis1977measurement}. The obtained agreement scores demonstrate substantial agreement across annotators for both evaluated models. Specifically, RoBERTa achieved a Fleiss’ Kappa score of $\kappa = 0.7947$ with an agreement percentage of 84.11\%, while DistilBERT achieved a Fleiss’ Kappa score of $\kappa = 0.7671$ with an agreement percentage of 82.24\%. Furthermore, the overall average Fleiss’ Kappa score across the evaluated models was $\kappa =0.7809$ with an average agreement percentage of 83.18\%, indicating strong consistency and reliability in the human annotation process. The human evaluation results further reveal that a considerable proportion of responses initially categorized as incorrect by automatic evaluation metrics were semantically meaningful according to human judgment. For RoBERTa, the final majority-vote annotation distribution indicated that 45.79\% of the responses were categorized as correct, 6.54\% as partially correct, and 47.66\% as incorrect. Similarly, for DistilBERT, 39.25\% of the responses were categorized as correct, 8.41\% as partially correct, and 52.34\% as incorrect. These observations suggest that strict lexical overlap-based metrics may underestimate the semantic validity of generated responses. Moreover, RoBERTa demonstrated comparatively stronger semantic alignment with the ground-truth responses than DistilBERT, as reflected by its higher proportion of semantically correct responses and lower proportion of incorrect predictions. Overall, these findings emphasize the importance of incorporating human-centered semantic evaluation alongside automatic metrics to obtain a more comprehensive assessment of question-answering systems under input perturbations.

\section{Error Analysis}
\label{err}
This study examines how models respond to controlled variations, with a specific focus on internal stochastic perturbations and modifications to input formulations. To provide a clearer understanding of how models behave under such conditions and to visually illustrate the potential pitfalls of unreliable behavior, this section presents illustrative examples from both datasets, where models exhibit hallucinated or inconsistent behavior when subjected to controlled perturbations. 
\par
One form of perturbation examined involves the application of Monte Carlo dropout, which introduces stochasticity during the inference process. By analyzing the model responses under internal variability, we assess their ability to produce stable outputs. Table \ref{ter} illustrates instances of model instability triggered by Monte Carlo dropout, underscoring the models' susceptibility to randomness in internal computation, where “--” denotes a blank or no response to the input query. 
\par
Another dimension of perturbation explored in this study involves the paraphrasing of input queries, which evaluates the extent to which models can generalize
\begin{table}[H]
 \centering
 \caption{Illustrations of the models' hallucination based on changes in their internal configurations.}
\resizebox{\textwidth}{!}{
\begin{tabular}{cccccc} \hline
Question& Ground Truth& RoBERTa & BERT-Base  & DistilBERT  & AlBERT         \\   \hline    
\makecell{What causes \\ Pauli repulsion?} & \makecell{ferimonic nature\\of electrons} & \makecell{ferimonic nature\\of electrons} &\makecell{What causes \\ Pauli repulsion?} & \makecell{ferimonic nature\\of electrons}& --   \\
\\
\makecell{What is the\\force that causes\\rigid strength \\in structures} & normal force   & --  & -- & \makecell{repulsive forces\\ of interaction\\ between atoms \\at close contact}& -- \\
\\
\makecell{In Newton's second\\law, what are\\the units of\\mass and force\\in relation\\of microscales?} & fixed & fixed & -- & \makecell{relative units\\of force \\and mass then\\are fixed}& -- \\
\\
\makecell{What made\\Ohio country\\vulnerable?}& \makecell{military roads\\to the area\\by Braddock\\and Frobes} & \makecell{constructions of\\military roads\\to the area\\by Braddock\\and Frobes}& \makecell{legal and\\illegal settlement}& military roads & \makecell{legal and\\illegal settlement} \\
\\
\makecell{When did\\British begin\\to build\\fort under\\William Trent?} & \makecell{early months\\of 1754} & 1754 & 17 & 1754 & -- \\  
\hline
\end{tabular} }
\label{ter}
\end{table}
\noindent
across linguistic variations and maintain consistent outputs. Ideally, a reliable model should remain robust when faced with minor lexical changes in input. However, evidence from Table \ref{se} reveals that, in some instances, models produce divergent or even misleading responses to the paraphrased queries. These inconsistencies highlight a reliance on surface-level phrasing rather than deeper semantic understanding, ultimately raising concerns regarding the reliability of such models in practical deployment scenarios.
\begin{table}[H]
    \centering
    \caption{Some Illustration of Model's Hallucination under Input Perturbation.}
    \resizebox{\textwidth}{!}{
    \begin{tabular}{ccccccc}  \hline
    \makecell{Original \\ Question} & \makecell{Paraphrased \\ Question} & \makecell{Ground \\ Truth} & RoBERTa & BERT-Base & DistilBERT & AlBERT   \\   \hline
\makecell{What \\describes the \\ proportionality\\ of acceleration\\ to force \\and mass?}& \makecell{What \\is the \\ proportionality\\ of acceleration\\ to force \\and mass?}& \makecell{Newton's \\ second \\law} & inverse & \makecell{inverse\\ proportionality\\ of acceleration\\ to force \\and mass} & inverse & \makecell{inverse\\ proportionality}\\
\\
\makecell{How many\\ troops were \\ defeated for\\ British in \\Battle of \\ Carillon ?} &\makecell{How many\\ troops were \\ defeated \\ in the \\Battle of \\ Carillon ?}& 18,000 & 3,600 & 3,600 & 3,600 & 18,000 \\
\\
\makecell{Given the \\ strength of \\ French forces \\at Louisberg, \\what did \\ Loudoun do ?}& \makecell{What did \\ Loudoun do \\given the \\ strength of \\ French forces \\at Louisberg ?} & \makecell{returned \\ to New \\ York }& \makecell{returned \\ to New \\ York } & \makecell{a massacre \\has occured \\at Fort \\ William Henry}& \makecell{returned \\ to New \\ York }& \makecell{Loudoun \\ returned \\ to New \\ York } \\
\\
\makecell{How were \\ British able \\ to cut \\ supplies to \\ Louisbourg ?} & \makecell{How did \\ 
 the British \\  cut off \\ supplies to \\ Louisbourg ?}& \makecell{deportation \\ of the \\French speaking \\Acadian \\ population \\ from the \\area}& \makecell{deportation \\ of the \\French speaking \\Acadian \\ population \\ from the \\area} & \makecell{deportation \\ of the \\French speaking \\Acadian \\ population \\ from the \\area} & \makecell{land-based\\ reinforcements } &\makecell{deportation \\ of the \\French speaking \\Acadian \\ population \\ from the \\area}  \\
\\
\makecell{Who did\\ Shirley leave\\ at Oswego ?}& \makecell{Who left\\ Shirley\\in Oswego ?}& garrisons & Johnson& Johnsono& the French & the French \\
\\   
 \hline
\end{tabular}
}
\label{se}
\end{table}
\noindent


\section{Discussion}
\label{dis}
A key observation is that even models with high accuracy can provide inconsistencies in their predictions, while lower-accuracy models may still demonstrate stability in outputs. The decoupling of accuracy and reliability suggests that a model’s trustworthiness should not be judged solely based on its accuracy but also on its ability to produce consistent responses across different dynamic conditions. This finding reinforces the importance of incorporating reliability assessments in model evaluations, particularly for applications where stability in predictions is critical. Additionally, it highlights the need to develop techniques that improve both accuracy and reliability to ensure more robust and dependable models.   \par
The study also emphasizes that dropout did not introduce any adverse changes to the model's behavior, indicating Monte Carlo dropout to be an effective approach for measuring reliability without disrupting model inference and reinforcing its utility for uncertainty estimation. Additionally, the ability of models to retain stability even under stochastic perturbations highlights the effectiveness of attention mechanisms, ensuring semantic understanding and reliability across dynamic conditions. This illustrates how a model's learning is resilient to changes in its internal configuration, emphasizing the capability of attention in maintaining effective learning. However, the impact of input perturbations reveals that even high-performing models struggle to maintain answer consistency under lexical changes. This suggests that model stability under lexical variation needs further improvement, particularly for conversational QA tasks, where variations in phrasing are common. This resilience is critical for real-world applications where consistent and accurate predictions are essential for tasks such as question answering and decision-making. Overall, these results contribute to the development of robust NLP systems capable of maintaining performance in diverse environments. \par
The study highlights the disadvantages of relying exclusively on accuracy as a performance metric. Although RoBERTa achieves high scores, its responses exhibit variability under controlled perturbations, underscoring the necessity for models that balance precision and stability. \par
Moreover, the study has several limitations. It relies solely on representation-based models and is limited to the question-answering task, although it could be extended to more complex NLP tasks. Additionally, it uses only two datasets, suggesting the need for broader evaluation. For variation generation, only one approach is used per type—paraphrasing for input variations and Monte Carlo dropout for internal model variations—whereas alternative methods could also be explored.

\section{Conclusion}
\label{con}
This research evaluates the reliability of BERT-based models on two QA tasks using a methodology that examines stability under internal configuration changes and input perturbations. Monte Carlo dropout was employed to introduce stochastic variations, while a pre-trained paraphrasing model was used to evaluate the impact of lexical modifications, providing a robust assessment of the models’ ability to maintain consistent outputs despite controlled variations. Experimental results show that RoBERTa, DistilBERT, and AlBERT outperform BERT Base, with performance varying across datasets. RoBERTa performs superior on both datasets in terms of accuracy and consistency, whereas DistilBERT is more stable in handling internal configuration variations, and ALBERT performs better when subjected to handling input variations, highlighting the scenario dependency. The statistical analyses further validate the existence of significant performance differences among models, reinforcing the importance of problem-specific as well as dataset-specific model selection rather than assuming universal effectiveness.  Moreover, the study highlights that accuracy alone is insufficient to represent a model's reliability and prediction stability, as input variations can significantly impact model consistency. \par 
This study provides a framework for evaluating the reliability of models, which is crucial for real-world applications such as virtual assistants, automated knowledge retrieval, and conversational AI systems. The insights gained can inform the development of more robust and stable QA models capable of handling variations in both data and model configurations. Future work could explore advanced uncertainty quantification techniques, such as Bayesian neural networks or ensemble learning, that further enhance model reliability and extend reliability assessments to more diverse and challenging problems. Additionally, domain-specific adaptations and fine-tuning strategies could be investigated to improve model stability across diverse linguistic contexts. These advancements would contribute to the development of more trustworthy and resilient NLP models for real-world deployment.


\section*{Data Availability}
The datasets used in this study are publicly available benchmark datasets. The SQuAD v2.0 dataset can be accessed at \href{https://rajpurkar.github.io/SQuAD-explorer/}{https://rajpurkar.github.io/SQuAD-explorer/}, and the QuAC dataset can be accessed at \href{https://quac.ai/}{https://quac.ai/}. The implementation code, experimental framework, and evaluation procedures used in this study are publicly available through an anonymous online repository for reproducibility purposes: \href{https://github.com/uncertainity-quantification/reliability-estimation-qa1}{https://github.com/uncertainity-quantification/reliability-estimation-qa1}. The repository contains the code and computational procedures required to reproduce the reported experimental results, statistical analyses, summary statistics, tables, and figures presented in this manuscript.

\section*{Disclosure of Interest} The authors declare that no conflict of interest could have influenced the work reported in this manuscript.

\section*{Funding Information}
The work was supported by the Slovene Research and Innovation Agency (ARIS) project GC-0002 and the core research programme P6-0411.  The work was also supported by EU through ERA Chair grant no. 101186647 (AI4DH).

\section*{Acknowledgments}
The authors gratefully acknowledge the collaboration and valuable contributions of the project partners and institutions involved in this research.

\newpage
\appendix
\section{Appendix: Comparative Results of LLM Evaluation}

To extend the scope of our reliability analysis beyond BERT-based architectures, we conducted additional experiments to examine the behaviour of Large Language Models (LLMs). The motivation for this was to determine whether LLMs, even in their compact forms, demonstrate higher consistency when confronted with semantically altered inputs. Recent research has also extensively explored the capabilities and limitations of large language models (LLMs), particularly in the context of question-answering (QA) tasks, including text generation and summarization. \cite{29} introduced GPT-3, a large-scale autoregressive generative model that demonstrates impressive text generation and few-shot learning abilities. However, despite its strong generative performance, GPT-3 showed inconsistent results on factual QA tasks, indicating limitations in precision and reliability. Further developments, such as PEGASUS by \cite{31} emphasized the generative capabilities of transformer-based models for abstractive summarization, confirming that generative architectures excel in text synthesis rather than factual consistency. More recent evaluations, including \cite{32}, empirically demonstrated that ChatGPT, while highly capable in open-ended generation, underperforms on structured QA benchmarks and often produces inconsistent or inaccurate responses. Similarly, \cite{33} surveyed hallucination issues in LLMs, reporting that these models frequently generate factually incorrect or fabricated content in QA settings. \par

As LLMs are often pre-trained on vast and diverse corpora, it is theoretically expected that they may exhibit more robust semantic understanding compared to earlier transformer-based models such as BERT, RoBERTa, or DistilBERT. However, large-scale models differ significantly in architecture, parameter scale, and training objectives. Therefore, evaluating a feasible, resource-efficient LLM, \textit{Tiny Llama}, allowed us to perform a practical comparison under similar experimental conditions, aligning with the reliability estimation framework proposed in the main text. \par

The results of these experiments are summarized in Tables \ref{cvb} and \ref{uhi}, which present the performance of Tiny Llama against stochastic variation and paraphrased input in both datasets. The tables report both Cosine Similarity and $F_1$ Score metrics for answerable and unanswerable questions. As shown in Tables \ref{cvb} and \ref{uhi}, the average cosine similarity and $F_1$ score for both types of questions are consistently below 0.5, indicating that Tiny Llama’s responses deviate substantially from the ground truth. This observation demonstrates that, despite the model’s large-scale pre-training and inherent generative capabilities, its performance on extractive QA tasks remains limited.

\begin{table}[htbp]
\centering
\caption{Comparative results of Tiny Llama accuracy and consistency against stochastic variation via Monte Carlo Dropout.}
\begin{tabular}{llcccc}
\hline
Datasets& Category & Acc  & MCD Acc.  & Cosine Simi. & F$_1$ Score  \\
       &   & (\%) & (\%) &  (avg ± std)  &(avg ± std)   \\
\hline
\multirow{3}{*}{SQuAD} & Ans & 2.6 & 3.6 & $0.4758 \pm 0.1208$ & $0.1896 \pm 0.1217$ \\ 
& No-Ans& 39.6 & 39.6 & $0.4098 \pm 0.01$ & $0.3960 \pm 0.01$ \\ 
& Total& 21.1 & 21.6 & $0.4428 \pm 0.0854$ & $0.2928 \pm 0.086$ \\ 
\\
\multirow{3}{*}{QuAC} & Ans & 0.1 & 0.1 & $0.3757 \pm 0.01$ & $0.1472 \pm 0.01$ \\ 
& No-Ans & 48.18 & 48.25 & $0.4921 \pm 0.01$ & $0.4618 \pm 0.01$ \\ 
& Total & 36.05 & 36.13 & $0.4378 \pm 0.01$ & $0.3048 \pm 0.01$ \\ 
\hline
\end{tabular}
\label{cvb}
\end{table}

\begin{table}[htbp]
\centering
\caption{Comparative results of Tiny Llama’s performance against paraphrased inputs.}
\begin{tabular}{llcccc}
\hline
Models & Category & \multicolumn{2}{c}{SQuAD} & \multicolumn{2}{c}{QuAC} \\  \hline
\multirow{2}{*}{} & \multirow{2}{*}{} & Cosine & F$_1$ & Cosine & F$_1$ \\
 &  & Similarity & Score & Similarity & Score \\
\hline
\multirow{3}{*}{Tiny Llama} & Ans & 0.3714 & 0.0878 & 0.5064 & 0.2004 \\ 
 & No-Ans & 0.0078 & 0.0007 & 0.1301 & 0.0181 \\ 
 & Total & 0.1838 & 0.0446 & 0.4408 & 0.1686 \\ 
\hline
\end{tabular}
\label{uhi}
\end{table}

The comparatively lower performance of Tiny Llama in these experiments aligns with the literature, which consistently shows that LLMs are primarily designed for text generation and summarization tasks, rather than extractive QA. In contrast, BERT-based models have achieved comparatively consistent performance on multiple QA benchmarks, making them more appropriate for reliability estimation in this domain. Given Tiny Llama’s low performance, attempting to assess its reliability on QA tasks would be of limited value, as unreliable outputs cannot meaningfully inform consistency metrics. On the other hand, evaluating the reliability of BERT variants is justified, as their high accuracy ensures that reliability estimates reflect the model’s behavior under controlled perturbations, rather than inherent performance deficiencies. \par

In conclusion, while generative LLMs excel in open-ended language modeling, their limitations in extractive QA tasks make them less suitable for reliability analysis in this context. The experiments presented here confirm that BERT-based models remain relevant candidates for investigating QA tasks, providing stable and accurate outputs that allow meaningful estimation of model stability. These findings underscore the importance of aligning model evaluation with task-specific capabilities, emphasizing that extractive models with comparatively consistent performance are crucial for assessing reliability across various fields.

\bibliography{references}

@article{1,
  title={Comparative Analysis of State-of-the-Art {Q\&A} Models: {BERT, RoBERTa, DistilBERT, and ALBERT on SQuAD v2} Dataset},
  author={{\"O}zkurt, Cem},
journal={ ADBA Chaos and Fractals},
    volume=1,number=1,pages={19-30},
    doi={https://doi.org/10.69882/adba.chf.2024073},
  year={2024}
}

@inproceedings{2,
  title={Comparative Analysis of Transformer based Models for Question Answering},
  author={Rawat, Anchal and Samant, Surender Singh},
  booktitle={2022 2nd International Conference on Innovative Sustainable Computational Technologies (CISCT)},
  pages={1--6},
  year={2022},
  xorganization={IEEE}
}

@article{3,
  author       = {Kate Pearce and
                  Tiffany Zhan and
                  Aneesh Komanduri and
                  Justin Zhan},
  title        = {A Comparative Study of Transformer-Based Language Models on Extractive
                  Question Answering},
  journal      = {CoRR},
  volume       = {abs/2110.03142},
  year         = {2021},
  url          = {https://arxiv.org/abs/2110.03142},
  eprinttype    = {arXiv},
  eprint       = {2110.03142},
  bibsource    = {dblp computer science bibliography, https://dblp.org}
}

@article{4,
  title={Transformer models used for text-based question answering systems},
  author={Nassiri, Khalid and Akhloufi, Moulay},
  journal={Applied Intelligence},
  volume={53},
  number={9},
  pages={10602--10635},
  year={2023},
  publisher={Springer}
}

@article{5,
  title={Attention is all you need},
  author={ Ashish Vaswani and Noam Shazeer and Niki Parmar and Jakob Uszkoreit and Llion Jones and Aidan Gomez and Łukasz Kaiser and Illia Polosukhin},
  journal={Advances in Neural Information Processing Systems},
  year={2017}
}

@inproceedings{6,
    title = "Assessing Factual Reliability of Large Language Model Knowledge",
    author = "Wang, Weixuan  and
      Haddow, Barry  and
      Birch, Alexandra  and
      Peng, Wei",
    xeditor = "Duh, Kevin  and
      Gomez, Helena  and
      Bethard, Steven",
    booktitle = "Proceedings of the 2024 Conference of the North American Chapter of the Association for Computational Linguistics: Human Language Technologies (Volume 1: Long Papers)",
    pages = "805--819",
}

@article{7,
  title={{To BAN or not to BAN: Bayesian attention networks for reliable hate speech detection}},
  author={Miok, Kristian and {\v{S}}krlj, Bla{\v{z}} and Zaharie, Daniela and Robnik-{\v{S}}ikonja, Marko},
  journal={Cognitive Computation},
  volume={14},
  number={1},
  pages={353--371},
  year={2022},
  publisher={Springer}
}

@article{8,
  title={Measuring reliability of large language models through semantic consistency},
  author={Raj, Harsh and Rosati, Domenic and Majumdar, Subhabrata},
  journal={arXiv preprint arXiv:2211.05853},
  year={2022}
}

@inproceedings{9,
  title={Prediction uncertainty estimation for hate speech classification},
  author={Miok, Kristian and Nguyen-Doan, Dong and {\v{S}}krlj, Bla{\v{z}} and Zaharie, Daniela and Robnik-{\v{S}}ikonja, Marko},
  booktitle={Statistical Language and Speech Processing: 7th International Conference, SLSP 2019, Proceedings 7},
  pages={286--298},
  year={2019},
}

@article{11,
  title={Semantic consistency for assuring reliability of large language models},
  author={Raj, Harsh and Gupta, Vipul and Rosati, Domenic and Majumdar, Subhabrata},
  journal={arXiv preprint arXiv:2308.09138},
  year={2023}
}

@inproceedings{12,
    title = "{SQ}u{AD}: 100,000+ Questions for Machine Comprehension of Text",
    author = "Rajpurkar, Pranav  and
      Zhang, Jian  and
      Lopyrev, Konstantin  and
      Liang, Percy",
    booktitle = "Proceedings of the 2016 Conference on Empirical Methods in Natural Language Processing",
    doi = "10.18653/v1/D16-1264",
    pages = "2383--2392"
}

@inproceedings{13,
    title = "{Q}u{AC}: Question Answering in Context",
    author = "Choi, Eunsol  and
      He, He  and
      Iyyer, Mohit  and
      Yatskar, Mark  and
      Yih, Wen-tau  and
      Choi, Yejin  and
      Liang, Percy  and
      Zettlemoyer, Luke",
    editor = "Riloff, Ellen  and
      Chiang, David  and
      Hockenmaier, Julia  and
      Tsujii, Jun{'}ichi",
    booktitle = "Proceedings of the 2018 Conference on Empirical Methods in Natural Language Processing",
    pages = "2174--2184",
}

@inproceedings{14,
  title={{How does BERT answer questions? A layer-wise analysis of transformer representations}},
  author={Van Aken, Betty and Winter, Benjamin and L{\"o}ser, Alexander and Gers, Felix A},
  booktitle={Proceedings of the 28th ACM international conference on information and knowledge management},
  pages={1823--1832},
  year={2019}
}

@article{15,
  title={{Natural language based analysis of SQuAD: An analytical approach for BERT}},
  author={Guven, Zekeriya Anil and Unalir, Murat Osman},
  journal={Expert Systems with Applications},
  volume={195},
  pages={116592},
  year={2022},
  publisher={Elsevier}
}

@article{16,
  author       = {Shilun Li and
                  Renee Li and
                  Veronica Peng},
  title        = {{Ensemble AlBERT on SQuAD 2.0}},
  journal      = {CoRR},
  volume       = {abs/2110.09665},
  year         = {2021},
  url          = {https://arxiv.org/abs/2110.09665},
  eprinttype    = {arXiv},
  eprint       = {2110.09665},
  bibsource    = {dblp computer science bibliography, https://dblp.org}
}

@article{17,
title = {A deep network model for paraphrase detection in short text messages},
journal = {Information Processing \& Management},
volume = {54},
number = {6},
pages = {922-937},
year = {2018},
xissn = {0306-4573},
doi = {https://doi.org/10.1016/j.ipm.2018.06.005},
xurl = {https://www.sciencedirect.com/science/article/pii/S0306457317308713},
author = {Basant Agarwal and Heri Ramampiaro and Helge Langseth and Massimiliano Ruocco}
}

@inproceedings{18,
  title={Paraphrasing for style},
  author={Xu, Wei and Ritter, Alan and Dolan, William B and Grishman, Ralph and Cherry, Colin},
  booktitle={Proceedings of COLING 2012},
  pages={2899--2914},
  year={2012}
}

@inproceedings{19,
  title={A review on different methods of paraphrasing},
  author={Gadag, Ashwini and Sagar, BM},
  booktitle={2016 International conference on electrical, electronics, communication, computer and optimization techniques (ICEECCOT)},
  pages={188--191},
  year={2016},
  xorganization={IEEE}
}

@article{20,
  title={Extractive text summarization},
  author={Mittal, Namita and Agarwal, Basant and Mantri, Himanshu and Goyal, Rahul Kumar and Jain, Manoj Kumar},
journal= {International Journal of Current  Engineering  and Technology},
volume=4,number=2,
  year={2014}
}

@article{24,
  title={Understanding dropout},
  author={Baldi, Pierre and Sadowski, Peter J},
  journal={Advances in neural information processing systems},
  volume={26},
  year={2013}
}

@inproceedings{26,
  title={{BERT: Pre-training} of deep bidirectional transformers for language understanding},
  author={Devlin, Jacob and Chang, Ming-Wei and Lee, Kenton and Toutanova, Kristina},
  booktitle={Proceedings of the 2019 conference of the North American chapter of the association for computational linguistics: human language technologies, volume 1 (long and short papers)},
  pages={4171--4186},
  year={2019}
}

@article{27,
  title={{RoBERTa: A} robustly optimized {BERT} pretraining approach},
  author={Liu, Yinhan and Ott, Myle and Goyal, Naman and Du, Jingfei and Joshi, Mandar and Chen, Danqi and Levy, Omer and Lewis, Mike and Zettlemoyer, Luke and Stoyanov, Veselin},
  journal={arXiv preprint arXiv:1907.11692},
  year={2019}
}

@article{28,
  title={{AlBERT: A} lite {BERT} for self-supervised learning of language representations},
  author={Lan, Zhenzhong and Chen, Mingda and Goodman, Sebastian and Gimpel, Kevin and Sharma, Piyush and Soricut, Radu},
  journal={International Conference on Learning Representations.},
  year={2020}
}

@article{29,
  title={Language models are few-shot learners},
  author={Brown, Tom and Mann, Benjamin and Ryder, Nick and Subbiah, Melanie and Kaplan, Jared D and Dhariwal, Prafulla and Neelakantan, Arvind and Shyam, Pranav and Sastry, Girish and Askell, Amanda and others},
  journal={Advances in neural information processing systems},
  volume={33},
  pages={1877--1901},
  year={2020}
}

@inproceedings{31,
  title={Pegasus: Pre-training with extracted gap-sentences for abstractive summarization},
  author={Zhang, Jingqing and Zhao, Yao and Saleh, Mohammad and Liu, Peter},
  booktitle={International conference on machine learning},
  pages={11328--11339},
  year={2020},
  organization={PMLR}
}

@article{32,
  title={Evaluating {ChatGPT} as a question answering system: {A} comprehensive analysis and comparison with existing models},
  author={Bahak, Hossein and Taheri, Farzaneh and Zojaji, Zahra and Kazemi, Arefeh},
  journal={arXiv preprint arXiv:2312.07592},
  year={2023}
}

@article{33,
  title={A survey on hallucination in large language models: Principles, taxonomy, challenges, and open questions},
  author={Huang, Lei and Yu, Weijiang and Ma, Weitao and Zhong, Weihong and Feng, Zhangyin and Wang, Haotian and Chen, Qianglong and Peng, Weihua and Feng, Xiaocheng and Qin, Bing and others},
  journal={ACM Transactions on Information Systems},
  volume={43},
  number={2},
  pages={1--55},
  year={2025},
  publisher={ACM New York, NY}
}

@article{ma2025dlcm,
  title={LE-DLCM: Decoupled Learner and Course Modeling with Large Language Models for Enhanced Course Recommendation},
  author={Ma, Jinjin and Zhao, Zhuo and Xie, Zhiwen and Zhang, Yi and Zhou, Guangyou},
  journal={Knowledge-Based Systems},
  pages={115135},
  year={2025},
  publisher={Elsevier}
}

@article{wu2024survey,
  title={A survey on large language models for recommendation},
  author={Wu, Likang and Zheng, Zhi and Qiu, Zhaopeng and Wang, Hao and Gu, Hongchao and Shen, Tingjia and Qin, Chuan and Zhu, Chen and Zhu, Hengshu and Liu, Qi and others},
  journal={World Wide Web},
  volume={27},
  number={5},
  pages={60},
  year={2024},
  publisher={Springer}
}

@article{lin2025can,
  title={How can recommender systems benefit from large language models: A survey},
  author={Lin, Jianghao and Dai, Xinyi and Xi, Yunjia and Liu, Weiwen and Chen, Bo and Zhang, Hao and Liu, Yong and Wu, Chuhan and Li, Xiangyang and Zhu, Chenxu and others},
  journal={ACM Transactions on Information Systems},
  volume={43},
  number={2},
  pages={1--47},
  year={2025},
  publisher={ACM New York, NY}
}

@article{shehmir2025llm4rec,
  title={LLM4Rec: a comprehensive survey on the integration of large language models in recommender systems—approaches, applications and challenges},
  author={Shehmir, Sarama and Kashef, Rasha},
  journal={Future Internet},
  volume={17},
  number={6},
  pages={252},
  year={2025},
  publisher={MDPI}
}

@article{landis1977measurement,
  title={The measurement of observer agreement for categorical data},
  author={Landis, J Richard and Koch, Gary G},
  journal={biometrics},
  pages={159--174},
  year={1977},
  publisher={JSTOR}
}

\end{document}